\documentclass{article}

\PassOptionsToPackage{numbers, compress}{natbib}

\usepackage{amssymb}
\usepackage{pifont}
\usepackage{makecell}
\usepackage{wrapfig}

\usepackage[eandd, final]{neurips_2026}

\usepackage[utf8]{inputenc} 
\usepackage[T1]{fontenc}    
\usepackage{hyperref}       
\usepackage{etoc}           
\usepackage{url}            
\usepackage{booktabs}       
\usepackage{amsfonts}       
\usepackage{nicefrac}       
\usepackage{tabularx}
\usepackage{microtype}      
\usepackage[table,xcdraw]{xcolor} 
\IfFileExists{xurl.sty}{\usepackage{xurl}}{}
\DeclareUrlCommand\mycode{\urlstyle{tt}}
\title{PertReason: A Knowledge-Grounded Benchmark and Framework for Cell-State-Conditioned Mechanistic Reasoning of Perturbation Effects}

\author{%
Dongkwan Kim\thanks{Equal contribution.} \quad Yiming Gao\footnotemark[1] \quad Yining Yang \quad Yang Shen
\\
Department of Electrical and Computer Engineering, Texas A\&M University \\
\texttt{\{dongkwan.kim, yiminggao618, yining\_yang, yshen\}@tamu.edu}
}

\usepackage[ruled,vlined]{algorithm2e}

\usepackage{graphicx} 
\usepackage{multirow}
\usepackage{enumitem}
\usepackage{CJKutf8}
\usepackage{xspace}
\usepackage{amssymb}

\usepackage{amsmath,amsfonts,bm}

\def\eqref#1{equation~\ref{#1}}

\def\1{\bm{1}}

\DeclareMathAlphabet{\mathsfit}{\encodingdefault}{\sfdefault}{m}{sl}
\SetMathAlphabet{\mathsfit}{bold}{\encodingdefault}{\sfdefault}{bx}{n}

\def\gG{{\mathcal{G}}}

\def\sA{{\mathbb{A}}}

\def\sS{{\mathbb{S}}}

\def\sV{{\mathbb{V}}}

\def\sZ{{\mathbb{Z}}}

\newsavebox{\shrinktablebox}
\newcommand{\shrinktable}[2]{%
  \sbox{\shrinktablebox}{#2}%
  \makebox[#1][c]{%
    \ifdim\wd\shrinktablebox>#1\relax
      \resizebox{#1}{!}{\usebox{\shrinktablebox}}%
    \else
      \usebox{\shrinktablebox}%
    \fi
  }%
}

\newcommand{\PertReasonQA}[0]{\textsc{PertReasonQA}\xspace}
\newcommand{\PertReasonLM}[0]{\textsc{PertReasonLM}\xspace}

\definecolor{weakgray}{HTML}{e7e7e7}
\definecolor{weakblue}{HTML}{E3F2FD}
\definecolor{weakred}{HTML}{FFEBEE}
\definecolor{midblue}{HTML}{bbdefb}
\definecolor{midred}{HTML}{ffcdd2}

\definecolor{darkblue}{HTML}{1565c0}
\definecolor{darkred}{HTML}{c62828}

\usepackage{spverbatim}
\usepackage{listings}
\usepackage{tcolorbox}
\usepackage{needspace}
\lstdefinestyle{wrappedtextbox}{
  basicstyle=\ttfamily\scriptsize,
  breaklines=true,
  breakatwhitespace=false,
  columns=fullflexible,
  keepspaces=true,
  showstringspaces=false,
  frame=single,
  framerule=0.3pt,
  rulecolor=\color{gray!60},
  backgroundcolor=\color{gray!5},
  aboveskip=4pt,
  belowskip=4pt
}

\begin{document}
\etocdepthtag.toc{main}

\maketitle

\begin{abstract}
Evaluating machine learning in scientific domains requires separating correct predictions from correct reasons under realistic distribution shifts. We introduce PertReason, a knowledge-grounded approach to evaluating and improving cell-state-conditioned reasoning about perturbation effects. Our framework comprises PertReasonQA, a benchmark that provides evaluation and supervision, and PertReasonLM, a reference large language model that demonstrates how such supervision supports reasoning-centric training. PertReasonQA tests whether models can generate mechanistically faithful explanations while remaining robust to complex shifts, such as new cells, unseen perturbations, cross-modal or combinatorial extrapolations. PertReasonQA combines single-cell genetic and chemical perturbation data across multiple cellular contexts with knowledge graphs, and dynamically conditions pathways on cell-specific basal states to avoid generic memorization. Evaluations of state-of-the-art models reveal systematic gaps between outcome prediction and mechanistic reasoning. As a reference probe of the benchmark, PertReasonLM is trained through supervised fine-tuning and reinforcement learning using mechanistic supervision from PertReasonQA. PertReasonLM improves predictive accuracy and reduces answer-reason disagreement by grounding rationales in context-specific pathways. Together, we provide a diagnostic framework for exposing and mitigating failures in faithful reasoning in data-rich scientific systems. We release our datasets to support further research at \url{https://dongkwan-kim.github.io/PertReason/}.
\end{abstract}

\section{Introduction}
Modeling complex and causal systems remains a fundamental challenge in machine learning. Virtual cell models serve as a demanding testbed for assessing AI reasoning and generalization within such environments, while accelerating drug discovery and disease modeling. Recent foundation models predict transcriptomic changes caused by chemical or genetic perturbations~\citep{cui2024scgpt, theodoris2023transfer, roohani2024predicting, wu2025perturbqa}. Despite their progress, two critical challenges limit their utility. First, these approaches suffer from a \textbf{lack of mechanistic reasoning}. Because they bypass established biological pathways to predict outcomes~\citep{novakovsky2023obtaining, dimitrov2026interpretation}, they cannot leverage shared causal mechanisms to generalize effectively across unseen perturbations~\citep{lotfollahi2023predicting, wei2025benchmarking, no_outperform_linear}. Second, \textbf{outcome-centric evaluation} protocols worsen this problem by judging models solely on their outcome predictions~\citep{wu2025perturbqa, op3, wei2025benchmarking, wu2025perturbench, wenteler2025pertevalscfm}. These benchmarks may fail to reveal whether models understand underlying biology or simply rely on spurious correlations.

To address these intertwined challenges, we introduce PertReason, which consists of a knowledge-grounded benchmark \PertReasonQA and a framework \PertReasonLM for rigorously evaluating and addressing mechanistic reasoning of perturbation effects, respectively. At its core, \PertReasonQA pairs cell-specific perturbation outcomes with outcome-compatible regulatory reference paths derived from knowledge graphs~\citep{turei2016omnipath, indra}. The path-construction procedure uses basal expression to weight candidate interactions, and a post hoc evidence audit characterizes the biological support and uncertainty of the resulting mechanistic proxies. This design separates correct predictions from correct reasons, exposing failures missed by standard outcome-centric benchmarks.

As a reference probe for this reasoning-centric benchmark, we present \PertReasonLM, a large language model trained to align outcome predictions with context-specific pathways. Existing approaches often rely on fixed embedding lookups~\citep{adduri2026predicting} or static graphs~\citep{roohani2024predicting, wenkel2025txpert}, which inherently struggle to generalize to new cellular contexts or new perturbations. \PertReasonLM instead acts as a semantic bridge that translates textual descriptions of novel experimental conditions into actionable regulatory states. By learning reasoning through Supervised Fine-Tuning (SFT) and Reinforcement Learning (RL), \PertReasonLM uses cell-conditioned pathway evidence to produce outcome-aligned causal explanations, reducing the failure modes exposed by \PertReasonQA.

We evaluate state-of-the-art models on \PertReasonQA across ten test folds covering new cellular contexts, unseen perturbations, cross-modal or combinatorial extrapolations. To move beyond outcome-centric metrics, we propose specialized evaluation protocols based on symbolic matching and functional similarity to assess the mechanistic faithfulness of the reasoning. Our experiments reveal a systematic gap invisible to outcome-centric benchmarks. Existing approaches, including Retrieval-Augmented Generation baselines~\citep{wu2025perturbqa} and general-purpose language models~\citep{yang2025qwen3technicalreport}, exhibit failure modes such as deriving correct answers through flawed logic or directionally inconsistent mechanisms. As a reference probe, \PertReasonLM tests whether reasoning-oriented training can close the gap between outcome prediction and mechanistic justification.

Our main contributions are summarized as follows. First, we introduce \PertReasonQA, a knowledge-grounded benchmark evaluating the mechanistic reasoning of perturbation effects using cell-conditioned pathways (\S\ref{sec:data}). Second, we present \PertReasonLM, a reference language model aligning outcome predictions with context-specific regulatory reasoning to address the failure modes (\S\ref{sec:model}). Third, using reasoning-centric evaluations, we expose systematic failures in existing models, such as flawed logic, ignored cellular contexts, and directionally inconsistent mechanisms (\S\ref{sec:results}).

\begin{figure*}[t]
\centering
\includegraphics[width=\textwidth]{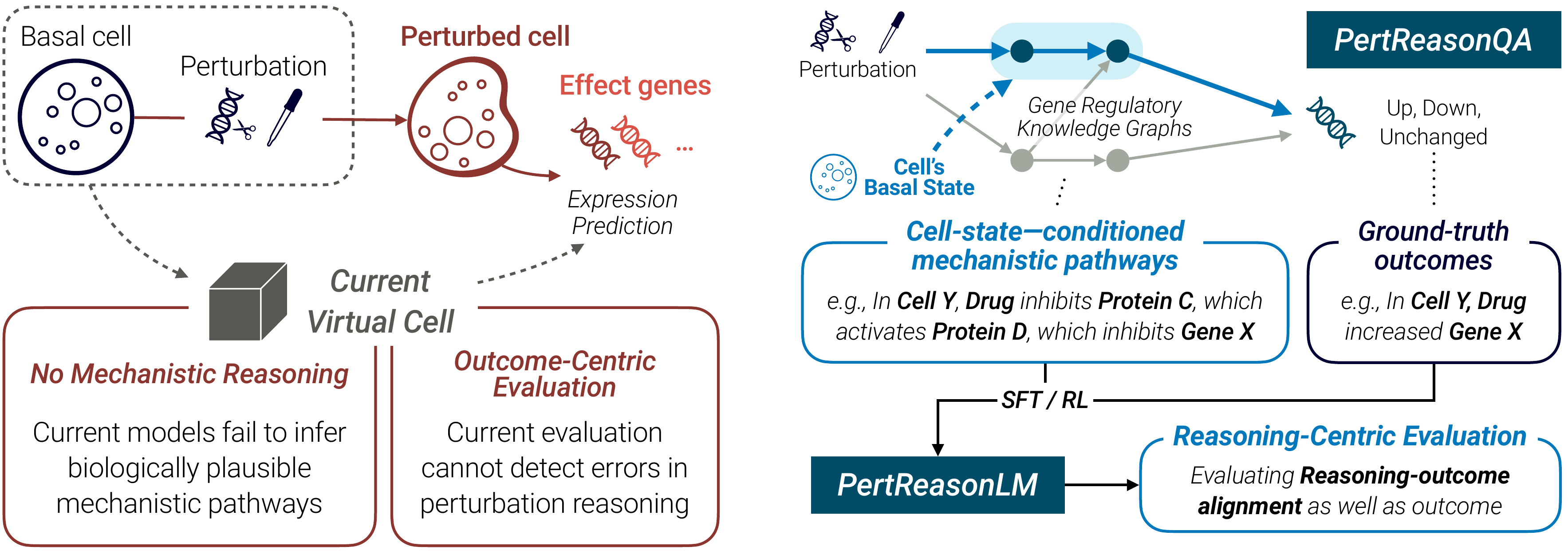}
\vspace{-0.48cm}
\caption{Overview of our proposed solutions in perturbation modeling. Current methods (left) predict expression changes without biological reasoning, limiting their evaluation to final outcomes. Our \PertReasonQA (right) leverages reasoning pathways from knowledge graphs, alongside \PertReasonLM, a reference baseline to validate the efficacy of the reasoning-centric benchmark.}
\label{fig:overview}
\vspace{-0.32cm}
\end{figure*}

\vspace{-0.17cm}
\section{Related Work}
\vspace{-0.15cm}

PertReason connects cellular perturbation prediction with knowledge-grounded biomedical reasoning. This section reviews both areas and their relationship to our benchmark and framework.

\vspace{-0.15cm}
\paragraph{Benchmarking and Modeling Cellular Perturbation Prediction}
Cellular perturbation prediction has advanced in parallel through modeling and benchmarking. Model architectures have evolved from early gene embeddings~\citep{du2019gene2vec} and generative models~\citep{lopez2018deep, gronbech2020scvae} to latent space arithmetic~\citep{lotfollahi2023predicting, hetzel2022predicting} and extensive single-cell foundation models~\citep{yang2022scbert, cui2024scgpt, theodoris2023transfer, rosen2023universal, adduri2026predicting}. Standardized benchmarks initially evaluated these systems through numerical regression of expression profiles~\citep{peidli2024scperturb, op3, wei2025benchmarking, wu2025perturbench, wenteler2025pertevalscfm}. However, optimizing strictly for regression yields black-box models that lack mechanistic reasoning, causing them to frequently underperform linear baselines~\citep{no_outperform_linear} or exhibit biological hallucinations~\citep{Radig2025hallucination}. To incorporate semantic context and structural grounding, recent modeling approaches have explored graph augmentations, synthetic trace imitation, and multi-agent pipelines~\citep{zhao2025biomazebenchmarkingenhancinglarge, phillips2025synthpertenhancingllmbiological, kim2026progressivemultiagentreasoningbiological}. Concurrently, PerturbQA~\citep{wu2025perturbqa} pioneered natural language evaluation, though its scope remains constrained to discrete factual question-answering rather than tracing complex causal mechanisms. \PertReasonQA addresses these gaps by shifting the evaluation paradigm from outcome predictions to the assessment of coherent, multi-step, and mechanistically faithful perturbation reasoning.

\vspace{-0.15cm}
\paragraph{Knowledge Graphs and Large Language Models in Biomedical Applications}

Researchers use graph-based methods, including Knowledge Graphs (KGs), to decipher complex biological interactions~\citep{chandak2023building, zitnik2018, kim2024translating, roohani2024predicting, ni2024identifying}. However, predefined KG schemas limit inductive generalization to novel entities~\citep{teru2020inductive}. Large Language Models (LLMs) bridge this semantic gap through biomedical reasoning~\citep{Luo_2022, Med-PaLM2} and linguistic representations of transcriptomics~\citep{singhal2022largelanguagemodelsencode, nori2023capabilitiesgpt4medicalchallenge, levine2024cell2sentence}. To ground LLMs in factual biology, recent frameworks integrate external knowledge via Retrieval-Augmented Generation (RAG)~\citep{lewis2021retrievalaugmentedgenerationknowledgeintensivenlp, wu2025perturbqa} or specialized agents~\citep{su2025kgarevion, gao2025scpilot}. Despite these synergies, current approaches prioritize discrete fact retrieval over dynamic causal simulation. \PertReasonQA evaluates mechanistically faithful reasoning beyond static predictions. \PertReasonLM aligns flexible language models with context-specific pathways, overcoming graph retrieval's structural limits.

\vspace{-0.17cm}
\section{PertReasonQA: Benchmarking of Perturbation Reasoning}\label{sec:data}
\vspace{-0.15cm}
In this section, we present \PertReasonQA, a novel question-answering (QA) benchmark designed to evaluate knowledge-grounded mechanistic reasoning about cellular responses to perturbations.

\vspace{-0.15cm}
\subsection{Problem Formulation and Benchmark Construction Pipeline}
\vspace{-0.09cm}

Cells respond to perturbations $p$ through intracellular signaling cascades that alter gene expression. A perturbation acts on a primary target, triggering regulatory events that ultimately affect a downstream effect gene $g_e$. Because these cascades strictly depend on the cell's basal state, mechanistic reasoning must identify the precise context-specific causal chain linking $p$ to the observed expression change.

\begin{figure*}[t]
\centering
\includegraphics[width=0.95\textwidth]{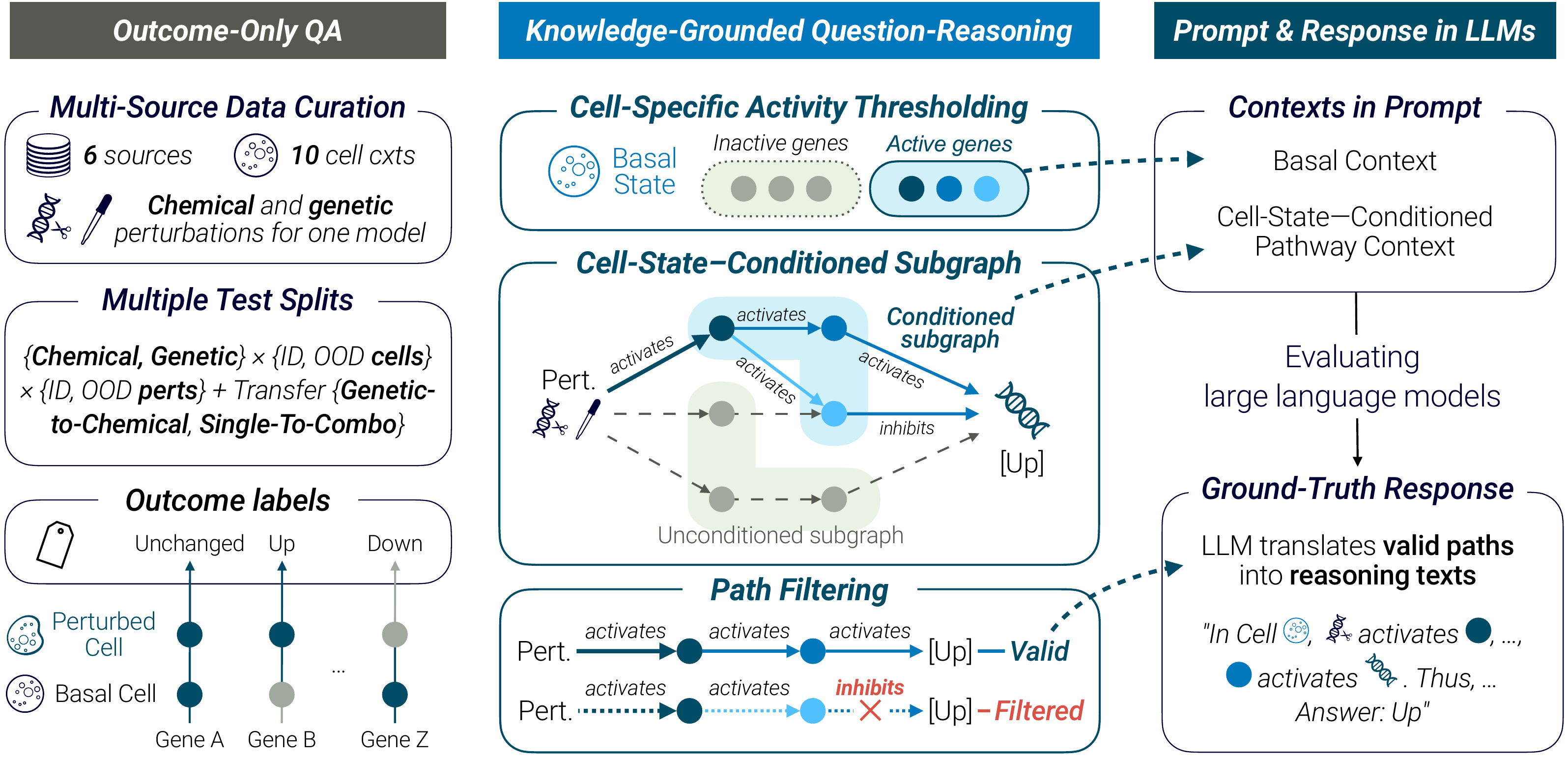}
\vspace{-0.02cm}
\caption{Overview of the \PertReasonQA construction pipeline. Multi-source data (left) cover chemical and genetic perturbations across ten test folds. Basal activity thresholds condition the knowledge graph on the cellular state, and extracted paths are filtered for consistency with observed outcomes (center). Basal and pathway contexts form the model prompts, while valid paths are translated into reference reasoning texts for evaluating mechanistic faithfulness (right).}
\vspace{-0.36cm}
\label{fig:data_pipeline}
\end{figure*}

We formalize \PertReasonQA as a knowledge-grounded QA task to assess this reasoning capacity. To build the benchmark, we implement a two-step data pipeline (Figure~\ref{fig:data_pipeline}). First, we construct outcome-only question-answering samples. Given a cellular context, a perturbation $p$, its directly perturbed gene set $\gG_0 = \{ g_0^{(1)},\ldots,g_0^{(m)} \}$, and an effect gene $g_e$, we extract the ground-truth differential expression labels $y \in \{\text{unchanged, up, down}\}$. Next, to move beyond outcome-centric evaluation, we synthesize knowledge-grounded question-reasoning samples. We pair these initial outcomes with reference reasoning paths $r$ that define the step-by-step biological mechanism as $[(p, e_0, g_0), \dots, (g_{k-1}, e_k, g_e)]$, where $g_0 \in \gG_0$. Here, each $e_i$ represents a regulatory interaction between neighboring genes derived from biological knowledge graphs.

\vspace{-0.15cm}
\subsection{Constructing Question-Outcome Dataset}

\vspace{-0.09cm}
\paragraph{Multi-Source Data Curation}

To convert high-dimensional single-cell profiles into discrete reasoning tasks for \PertReasonQA, we first derive outcome labels from chemical and genetic perturbation data from six large-scale sources~\citep{sciplex3, op3, Replogle, Nadig, demeo2025active, norman2019exploring}. Our dataset comprehensively spans ten cellular contexts comprising six cell lines (K562, MCF7, A549, HepG2, Jurkat, and RPE1) and four immune-cell types (B, Myeloid, NK, and T). Individual cells are aggregated into pseudobulks and paired with their respective average basal expression profiles.

\vspace{-0.15cm}
\paragraph{Splits for Rigorous Evaluation}

We rigorously evaluate out-of-distribution (OOD) generalization on ten test folds. Eight cross in-distribution (ID) and OOD cellular contexts with ID/OOD perturbations for each modality. The other two test genetic-to-chemical transfer within HepG2 and K562 CRISPRa single-to-combo transfer after training on single perturbations from other sources. We employ a network-based data split designed to limit memorization of overlapping pathways. Genetic ID/OOD assignments follow clusters of gene network embeddings from OmniPath~\citep{turei2016omnipath} and MSigDB~\citep{liberzon2015molecular}; drugs mapped via INDRA CoGEx~\citep{indra} are OOD if $>10\%$ of targets are in OOD clusters (otherwise ID). This structural design probes mechanistic extrapolation under functionally defined perturbation shifts (details in Appendix~\ref{app:splits}).

\vspace{-0.15cm}
\paragraph{Differential Expression Labeling}

Finally, we derive ground-truth outcome labels through differential expression analysis against control groups. Following PerturbQA~\citep{wu2025perturbqa}, we use log-transformed expression within each cell context and apply the Wilcoxon rank-sum test~\citep{wilcoxon1945individual} with Benjamini-Hochberg correction~\citep{benjamini2000adaptive}. We discretize continuous expression changes into three classes: Up-regulation or Down-regulation for genes with an adjusted $p$-value $<0.01$, and Unchanged for genes with a $p$-value $>0.1$. Genes in the intermediate interval $0.01 \le p \le 0.1$ are excluded from benchmark construction.

\vspace{-0.15cm}
\subsection{Synthesizing Knowledge-Grounded Reasoning Dataset}
\vspace{-0.09cm}

We synthesize a large-scale dataset of reasoning paths by mapping empirical transcriptomic observations to knowledge graphs. This process involves defining basal cellular states, tracing causal interactions from the perturbation to the effect gene, and converting these mechanisms into natural language. Detailed descriptions with rigorous notations are provided in the Appendix~\ref{app:reasoning}.

\vspace{-0.15cm}
\paragraph{Cell-Specific Basal Activity Thresholding}
To enable large language models to incorporate cellular context into semantic logic rather than processing raw numerical values, we discretize basal expression into interpretable biological states: ``low," ``medium," and ``high." To achieve this while effectively mitigating technical noise and batch effects, we establish adaptive, cell-specific thresholds using Kolmogorov-Smirnov (KS) statistics. Specifically, we define a low threshold $\tau_{\mathrm{low}}$ by applying a two-sample KS test to distinguish the empirical groups of genes labeled unchanged and changed. We then determine a high threshold $\tau_{\mathrm{high}}$ using a one-sample KS test to identify the saturation point of functionally active genes. We calibrate these thresholds using differential-expression labels only within the training set; at test time, they are reused for seen cell contexts or predicted for unseen contexts solely from basal expression distributions. This process converts numerical profiles into qualitative states, providing a robust foundation for mechanistic reasoning in \PertReasonQA.

\vspace{-0.15cm}
\paragraph{Cell-State-Conditioned Subgraph Extraction}
General KGs assemble regulatory interactions across diverse cellular states, providing a broad signaling landscape that requires contextualization for a given transcriptomic state. We dynamically refine the static graph into a context-specific subgraph by modulating edge weights based on the cell's basal expression $\mathbf{b}^{(c)}$. For each edge $(u, v)$, we define the conditioned weight as $w^{(c)}_{uv} = w^{\mathrm{base}}_{uv} \cdot \gamma^{\mathrm{src}}_u \cdot \gamma^{\mathrm{tgt}}_v$, where $\gamma_{\cdot} \propto \textrm{sigmoid}(\mathbf{b}_{\cdot}^{(c)} - \tau_{\mathrm{low}})$ is a sigmoid gate penalizing edges involving low-expression genes, where $\tau_{\mathrm{low}}$ is the cell-specific activity threshold. Based on this, we extend TieDIE~\citep{tiedie} to identify mechanistic paths from the perturbation to the effect gene. Using repeated Dijkstra searches on the weighted KG with perturbation-centered Personalized PageRank (PPR) and edge penalization, we extract a diverse ensemble of high-confidence causal paths. Basal expression provides contextual evidence for ranking paths by basal-state compatibility, while nonzero source and target floors preserve traversal through low-expression genes.

\vspace{-0.15cm}
\paragraph{Reference Paths via Evidence Augmentation and Filtering} 

To ensure the quality of reference reasoning, we filter the extracted paths to retain only those that logically justify the observed transcriptomic outcomes. We first resolve regulatory directions by assigning activating ($+1$) or inhibitory ($-1$) signs to edges based on consensus from INDRA CoGEx~\citep{indra}. For differentially expressed genes, we select only those paths where the cumulative sign propagation matches the observed direction of change (Algorithm~\ref{alg:reachability}). For unchanged responses, we identify cases where regulatory signals are masked by cellular boundary conditions (Algorithm~\ref{alg:consistency}), such as inhibitory signals targeting already silent genes. We assess biological support separately through a post hoc audit of frozen references (Appendix~\ref{app:reference_audit}). 

\vspace{-0.15cm}
\paragraph{Path-to-Text Reasoning Synthesis}

To enable language models to natively process biologically grounded mechanisms, we convert the fixed, filtered causal paths into natural language. We employ Qwen3-4B~\citep{yang2025qwen3technicalreport} to translate these paths into cohesive narrative explanations. Finally, we apply a post-generation keyword filter for explicit contradiction or non-support statements. Note that this filter does not independently establish biological validity or detect all hallucinations.

\vspace{-0.15cm}
\paragraph{Dataset Statistics} 

\PertReasonQA contains 915K outcome-only samples for training and validation, and 171K reasoning samples across training, validation, and test evaluation. Test reasoning samples are constructed from their held-out perturbation outcomes. Table~\ref{tab:data_stats} summarizes reasoning-sample statistics, with full statistics in Table~\ref{tab:data_stats_full} (Appendix~\ref{app:data}).

\renewcommand{\arraystretch}{1.1}
\begin{table}[t]
\caption{Summary statistics of reasoning samples in \PertReasonQA across training, validation, and ten test folds. Full statistics including outcome-only samples are in Table~\ref{tab:data_stats_full}.}
\vspace{-0.02cm}
\label{tab:data_stats}
\setlength{\tabcolsep}{2pt}
\shrinktable{\textwidth}{%
\begin{tabular}{lrrrrrrrrrrrrrr}
\toprule
{} & \multicolumn{2}{c}{Train ID Cell} & \multicolumn{2}{c}{Valid ID Cell} & \multicolumn{4}{c}{Test ID Cell} & \multicolumn{4}{c}{Test OOD Cell} & \multicolumn{1}{c}{\multirow{2}{*}{\begin{tabular}[c]{@{}c@{}}Gen-to\\ Chem\end{tabular}}} & \multicolumn{1}{c}{\multirow{2}{*}{\begin{tabular}[c]{@{}c@{}}Single-to\\ Combo\end{tabular}}} \\ \cmidrule(lr){2-13}
{} & \multicolumn{1}{c}{Chem.} & \multicolumn{1}{c}{Gen.} & \multicolumn{1}{c}{Chem.} & \multicolumn{1}{c}{Gen.} & \multicolumn{1}{c}{ID Chem.} & \multicolumn{1}{c}{ID Gen.} & \multicolumn{1}{c}{OOD Chem.} & \multicolumn{1}{c}{OOD Gen.} & \multicolumn{1}{c}{ID Chem.} & \multicolumn{1}{c}{ID Gen.} & \multicolumn{1}{c}{OOD Chem.} & \multicolumn{1}{c}{OOD Gen.} & & \\ \midrule
Cellular Contexts & 5 & 3 & 5 & 3 & 5 & 3 & 5 & 3 & 2 & 1 & 2 & 1 & 1 & 1 \\ 
Unique Perturbations & 127 & 636 & 17 & 69 & 14 & 67 & 21 & 15 & 14 & 62 & 19 & 12 & 4 & 5 \\ \midrule
Unchanged & 19111 & 20123 & 2678 & 2347 & 2316 & 1275 & 1913 & 62 & 809 & 192 & 815 & 24 & 1788 & 543 \\
Up & 17730 & 13000 & 1327 & 2357 & 1579 & 1265 & 628 & 133 & 803 & 396 & 496 & 64 & 388 & 147 \\
Down & 29122 & 26817 & 3149 & 4085 & 3252 & 1939 & 3203 & 464 & 1406 & 692 & 1935 & 273 & 467 & 210 \\ \midrule
Avg. Pathway Length & 5.98 & 4.55 & 6.11 & 4.22 & 6.25 & 4.68 & 5.29 & 3.95 & 6.05 & 4.89 & 5.07 & 4.35 & 6.05 & 6.01 \\ \bottomrule
\end{tabular}%
}
\vspace{-0.32cm}
\end{table}

\vspace{-0.15cm}
\subsection{Evaluation Metrics}\label{sec:evaluation}
\vspace{-0.09cm}

We evaluate models along two complementary axes: outcome accuracy and mechanistic faithfulness. 

\vspace{-0.15cm}
\paragraph{Outcome Accuracy}
We evaluate whether the model predicts the correct final perturbation outcome. 
We extract the final prediction from the last valid \texttt{<answer>} block and compare it against the ground-truth differential-expression label. 
We report balanced accuracy for outcome labels, \textit{up}, \textit{down}, and \textit{unchanged}, because label distributions vary across perturbation types, cell contexts, and test folds.

\vspace{-0.15cm}
\paragraph{Mechanistic Faithfulness}
Outcome accuracy alone cannot determine whether a model reaches the correct answer through a reference-consistent biological mechanism. 
We canonicalize the regulatory triplets extracted from the \texttt{<triplet>} block and evaluate the resulting structured mechanism.

\vspace{-0.15cm}
\begin{itemize}[leftmargin=1.0em]
    \item \textbf{Edge Recall:} 
    We compare the predicted triplets against the curated reference paths and measure the fraction of reference edges recovered by the model. We emphasize recall because language models may generate additional regulatory nodes or edges that are biologically plausible but absent from the curated reference path. We complement it with edge precision and F1 (Appendix~\ref{app:metric_diagnostics}).

    \item \textbf{Path Connectivity:} 
    Edge-level overlap does not guarantee that the predicted mechanism forms a coherent causal route. 
    We therefore verify whether the generated triplets form a connected chain from the perturbation source to the queried effect gene. 
    This metric detects fragmented mechanisms with individual plausible edges but no end-to-end perturbation-effect explanation.

    \item \textbf{Gene-Ontology Similarity:} 
    We measure broad functional relatedness between predicted and reference paths through Gene Ontology-Biological Process similarity (GO-BP). For each path, we collect the intermediate genes, exclude the perturbation and effect gene, construct IDF-weighted GO-BP profiles, and compute cosine similarity between the predicted and reference profiles.
\end{itemize}
\vspace{-0.09cm}

To assess outcome accuracy and mechanistic faithfulness jointly, we report a four-level answer-reason taxonomy based on final-answer correctness and mechanistic correctness. A generated reason is counted as correct when triplets exist, they form a directed connected path, and edge recall is at least $0.5$. In addition, supplementary trace diagnostics and controlled stress tests comparing original and systematically modified reference paths are detailed in Appendix~\ref{app:metric_diagnostics}. To complement these automatic measures, Appendix~\ref{app:trace_audit} examines their relationship to source-blinded biological support judgments.

\vspace{-0.17cm}
\section{Models and Experimental Setup}\label{sec:model}
\vspace{-0.15cm}

To investigate the effectiveness of reasoning-centric modeling, we introduce \PertReasonLM, the reference probe based on large language models. Our \PertReasonLM is trained to reason perturbation effects via a comprehensive pipeline of supervised fine-tuning (SFT) and group relative policy optimization (GRPO). We also describe the experimental configurations for our benchmarking.

\vspace{-0.15cm}
\subsection{\PertReasonLM: Perturbation Reasoning Language Model via SFT and GRPO}
\vspace{-0.09cm}

\paragraph{Supervised Fine-Tuning}

The initial SFT stage instills foundational biological knowledge and structural reasoning into \PertReasonLM via a multi-stage curriculum. First, to overcome the scarcity of high-quality reasoning paths, we employ outcome-only supervision. By predicting final differential expression labels directly from cellular contexts and perturbations, the model acquires broad domain priors and stable entity representations. We then introduce mechanistic reasoning supervision to align these learned associations with explicit causal mechanisms. \PertReasonLM is trained to generate natural language Chain-of-Thought (CoT) reasoning~\citep{wei2022chain, chung2024scaling, magister2023teaching} that justifies the final outcome using context-specific regulatory pathways. We format target responses with specialized tags (e.g., \texttt{<thinking>}, \texttt{<answer>}, \texttt{<triplet>}) and employ data mixing~\citep{guo2025deepseek, mitra2023orca} to integrate complex reasoning samples with outcome-only data, ensuring predictive stability. Finally, to handle ambiguous gene interactions, we integrate a relation direction self-supervision objective. By masking and predicting the causal signs of known edges~\citep{teru2020inductive, zhang2024making}, the model actively uses cellular and pathway contexts to learn edge-level mechanistic primitives (i.e., activation or inhibition). Our curriculum combines abundant outcome labels with scarce high-fidelity reasoning signals to develop a cohesive perturbation reasoning capability. SFT details are in Appendix~\ref{app:sft_impl}.

\vspace{-0.15cm}
\paragraph{Reinforcement Learning with GRPO}

We apply Group Relative Policy Optimization (GRPO)~\citep{shao2024deepseekmath} on top of the SFT model as a post-training alignment stage. It aims to tighten the agreement between the final expression label and the generated causal chain. To do so, the reward combines answer correctness with structured triplet-level feedback, so that the model is encouraged to produce responses that are both label-correct and mechanistically grounded.
Full GRPO implementation details and reward definitions are deferred to Appendix~\ref{app:rl_impl}.

\vspace{-0.15cm}
\paragraph{Backbone Models}

We use Qwen3-4B~\citep{yang2025qwen3technicalreport} as the primary backbone for \PertReasonLM due to its robust reasoning performance at a moderate parameter scale. We further conduct ablation experiments using the Qwen3-1.7B and Qwen3-8B. 

\vspace{-0.15cm}
\subsection{Pathway Contexts in Inference Stages}
\vspace{-0.09cm}

To examine how language models use pathway information, we define four inference scenarios. First, \textbf{no path} is a baseline without pathway input. Second, \textbf{unconditional path} provides regulatory chains from the unweighted global KG, independent of cell states. Third, the \textbf{cell-conditioned path} is our \textit{default}, with pathways dynamically selected based on the cell-specific basal state; these paths are contextually grounded but remain imperfect because their potential regulatory trajectories may not directly support the observed outcome. Finally, \textbf{reference path} consists of reference pathways mechanistically consistent with the observed outcome. These provide a performance ceiling but are unavailable for real-world inference since they require access to the ground-truth outcomes.

\vspace{-0.15cm}
\subsection{Baselines}
\vspace{-0.09cm}

We compare \PertReasonLM with state-of-the-art gene-space and text-space baselines. Baseline descriptions, evaluation protocols, and configurations are detailed in Appendix~\ref{app:baselines}.

\vspace{-0.15cm}
\paragraph{Gene-Space Baselines with Numerical Outputs}

We adopt: (1) GEARS~\cite{roohani2024predicting}, which utilizes a gene-gene similarity graph to model genetic perturbation effects, (2) scGPT~\cite{cui2024scgpt}, and (3) STATE (SE+ST)~\cite{adduri2026predicting}, generative foundation models pre-trained on single-cell transcriptomic data. We evaluate these baselines only on genetic perturbations because GEARS and scGPT are architecturally limited to them, while STATE's chemical embedding lookup does not support OOD generalization. All three are trained on CRISPRi data and evaluated only on the four matched CRISPRi genetic folds, excluding CRISPRa combinations. Chemical proxy limitations are detailed in Appendix~\ref{app:chemical_baseline_limitations}.

\vspace{-0.15cm}
\paragraph{Text-Space Baselines Based on LLMs}
We compare: (1) the {Qwen-3 4B} base model~\cite{yang2025qwen3technicalreport}, (2) domain-specific LLMs, {BioMistral 7B}~\cite{labrak2024biomistralcollectionopensourcepretrained} and {NatureLM 8x7B}~\cite{xia2025naturelanguagemodeldeciphering}, (3) {SUMMER}~\cite{wu2025perturbqa} with a Qwen3-4B backbone, a Retrieval-Augmented Generation (RAG) model for perturbation responses.

\vspace{-0.15cm}
\paragraph{Proprietary LLM APIs}
We also test the proprietary models GPT-5.4 and GPT-5.6 Sol. To assess the performance of resource-intensive API-based models, we create \PertReasonQA-mini, a subset of the ten test folds comprising a reported 1,991 examples, with each test slice capped at 200 instances. As detailed in Appendix~\ref{app:mini_benchmark}, this mini-benchmark serves as a proxy for the full suite.

\vspace{-0.17cm}
\section{Results and Discussion}\label{sec:results}
\vspace{-0.15cm}

We first compare outcome prediction on \PertReasonQA across in-domain and out-of-distribution settings, then examine whether correct predictions are supported by faithful cell-conditioned mechanisms. We further analyze the effects of pathway context, perturbation modality, and model scale to identify where mechanistic supervision improves generalization and where current models still fail.

\renewcommand{\arraystretch}{1.1}
\begin{table}[t]
\caption{Balanced accuracy for outcome prediction on 3-way gene expression QA across 10 test folds. Results are reported under the cell-conditioned path context (default). Averages are computed over the displayed slices; GEARS, scGPT, and STATE are trained and evaluated only on the genetic CRISPRi subset by design (see \S\ref{app:baselines_gene}).}
\vspace{-0.09cm}
\label{tab:main_path}
\setlength{\tabcolsep}{2.5pt}
\shrinktable{\textwidth}{%
\begin{tabular}{@{}lccccccccccc@{}}
\toprule
                        &                               & \multicolumn{4}{c}{ID Cell}                                                                                                   & \multicolumn{4}{c}{OOD Cell} & & \\ \cmidrule(lr){3-10}
\multirow{-2}{*}{Model} & \multirow{-2}{*}{Average}     & ID Chem.                      & ID Gen.                       & OOD Chem.                     & OOD Gen.                      & ID Chem.                      & ID Gen.                       & OOD Chem.                     & OOD Gen. & \multirow{-2}{*}{\begin{tabular}[c]{@{}c@{}}Gen-to\\ Chem\end{tabular}} & \multirow{-2}{*}{\begin{tabular}[c]{@{}c@{}}Single-to\\ Combo\end{tabular}} \\ \midrule
GEARS                   & \cellcolor[HTML]{EFB3B3}0.356 & -                             & \cellcolor[HTML]{EFB3B3}0.357 & -                             & \cellcolor[HTML]{F4C7C7}0.405 & -                             & \cellcolor[HTML]{EEADAD}0.341 & -                             & \cellcolor[HTML]{ECA4A4}0.320 & - & - \\
scGPT                   & 0.359 & -                             & 0.374 & -                             & 0.390 & -                             & 0.350 & -                             & 0.323 & - & - \\
STATE                   & \cellcolor[HTML]{F4CACA}0.411 & -                             & \cellcolor[HTML]{FCF3F3}0.510 & -                             & \cellcolor[HTML]{F4C8C8}0.407 & -                             & \cellcolor[HTML]{F3C7C7}0.403 & -                             & \cellcolor[HTML]{EDA6A6}0.324 & - & - \\ \midrule
BioMistral 7B & {0.313} & \cellcolor[HTML]{EB9E9E}0.305 & \cellcolor[HTML]{EB9F9F}0.307 & \cellcolor[HTML]{ECA1A1}0.313 & \cellcolor[HTML]{ECA4A4}0.319 & \cellcolor[HTML]{EDA6A6}0.324 & \cellcolor[HTML]{EA9999}0.294 & \cellcolor[HTML]{ECA2A2}0.315 & \cellcolor[HTML]{EA9999}0.293 & \cellcolor[HTML]{EEACAC}0.340 & {0.323} \\
NatureLM 8x7B & {0.334} & \cellcolor[HTML]{EEABAB}0.336 & \cellcolor[HTML]{EDA9A9}0.332 & \cellcolor[HTML]{EDAAAA}0.334 & \cellcolor[HTML]{EDAAAA}0.333 & \cellcolor[HTML]{EDA9A9}0.331 & \cellcolor[HTML]{EEABAB}0.337 & \cellcolor[HTML]{EEAAAA}0.335 & \cellcolor[HTML]{EDAAAA}0.333 & \cellcolor[HTML]{EDAAAA}0.334 & {0.331} \\
SUMMER-4B & {0.397} & \cellcolor[HTML]{F7D7D7}0.444 & \cellcolor[HTML]{F2C1C1}0.389 & \cellcolor[HTML]{F5CECE}0.421 & \cellcolor[HTML]{F0B5B5}0.360 & \cellcolor[HTML]{F7D7D7}0.442 & \cellcolor[HTML]{F4CCCC}0.416 & \cellcolor[HTML]{F2C2C2}0.392 & \cellcolor[HTML]{F0B8B8}0.367 & \cellcolor[HTML]{F2BEBE}0.382 & {0.359} \\ \midrule
Qwen3-4B 0-shot & {0.364} & \cellcolor[HTML]{F5CDCD}0.418 & \cellcolor[HTML]{EDA8A8}0.330 & \cellcolor[HTML]{F0B4B4}0.358 & \cellcolor[HTML]{F0B4B4}0.358 & \cellcolor[HTML]{F3C5C5}0.400 & \cellcolor[HTML]{F1BABA}0.372 & \cellcolor[HTML]{EDA6A6}0.325 & \cellcolor[HTML]{F1B9B9}0.370 & \cellcolor[HTML]{EEADAD}0.341 & {0.372} \\ \midrule
\PertReasonLM-4B           & \multicolumn{1}{l}{}          & \multicolumn{1}{l}{}          & \multicolumn{1}{l}{}          & \multicolumn{1}{l}{}          & \multicolumn{1}{l}{}          & \multicolumn{1}{l}{}          & \multicolumn{1}{l}{}          & \multicolumn{1}{l}{}          & \multicolumn{1}{l}{} & \multicolumn{1}{l}{} & \multicolumn{1}{l}{} \\
\quad Outcome SFT & \cellcolor[HTML]{E4EDFC}0.613 & \cellcolor[HTML]{BFD4F7}0.713 & \cellcolor[HTML]{C7DAF8}0.690 & \cellcolor[HTML]{E4EDFC}0.612 & \cellcolor[HTML]{E1EBFB}0.622 & \cellcolor[HTML]{D5E3FA}0.653 & \cellcolor[HTML]{D8E5FA}0.646 & \cellcolor[HTML]{FEFEFF}0.543 & \cellcolor[HTML]{E5EDFC}0.611 & \cellcolor[HTML]{FAE7E7}0.482 & \cellcolor[HTML]{FAFBFE}0.554 \\
\quad + CoT SFT & \cellcolor[HTML]{CBDCF9}0.681 & \cellcolor[HTML]{B2CBF6}0.749 & \cellcolor[HTML]{BCD2F7}0.722 & \cellcolor[HTML]{CCDDF9}0.678 & \cellcolor[HTML]{C1D5F7}0.708 & \cellcolor[HTML]{C2D6F8}0.705 & \cellcolor[HTML]{C5D8F8}0.697 & \cellcolor[HTML]{CFDFF9}0.669 & \cellcolor[HTML]{B4CDF6}0.743 & \cellcolor[HTML]{E6EEFC}0.608 & \cellcolor[HTML]{FFFDFD}0.535 \\
\quad + GRPO & \cellcolor[HTML]{C2D6F8}0.704 & \cellcolor[HTML]{A9C5F5}0.772 & \cellcolor[HTML]{AFC9F5}0.756 & \cellcolor[HTML]{D0DFF9}0.668 & \cellcolor[HTML]{ACC8F5}0.763 & \cellcolor[HTML]{BCD2F7}0.721 & \cellcolor[HTML]{B3CCF6}0.745 & \cellcolor[HTML]{CBDCF9}0.680 & \cellcolor[HTML]{A4C2F4}0.786 & \cellcolor[HTML]{E5EEFC}0.609 & \cellcolor[HTML]{FFFEFE}0.537 \\ \bottomrule
\end{tabular}%
}
\vspace{-0.35cm}
\end{table}

\vspace{-0.15cm}
\subsection{Benchmarking Outcome Prediction and Generalization}
\vspace{-0.09cm}

We evaluate model performance under the cell-conditioned path context, where models must reason through noisy, context-specific regulatory environments. Table~\ref{tab:main_path} summarizes the balanced accuracy across ten test folds. Gene-space baselines, including GEARS, scGPT, and STATE, generally perform worse in unseen cellular or perturbation contexts. While STATE achieves a peak balanced accuracy of 0.510 on in-domain data, its performance drops substantially in OOD settings, suggesting limited transfer of learned perturbation--response relationships to unseen contexts. Similarly, evaluations on the \PertReasonQA-mini subset (Appendix~\ref{app:mini_benchmark}) reveal that GPT-5.4 and GPT-5.6 Sol achieve balanced accuracies of 0.386 and 0.437, respectively, above Qwen3-4B (0.371) but below the trained \PertReasonLM models.

\PertReasonLM demonstrates superior predictive power and robustness against distribution shifts. While the \PertReasonLM-SFT (Outcome) variant achieves an average balanced accuracy of 0.613, incorporating chain-of-thought reasoning in \PertReasonLM-SFT pushes performance to 0.681. \PertReasonLM-GRPO further improves this to 0.704, outperforming SFT in nine out of ten test folds and suggesting that GRPO helps align final outcome predictions with the generated mechanisms. \PertReasonLM-GRPO shows little degradation from ID to OOD settings, whereas BioMistral and NatureLM struggled with the reasoning protocol (see Appendix~\ref{app:biomed_llm_failures} for common failure modes).

Across evaluation settings, \PertReasonLM generally maintains higher outcome accuracy than the evaluated baselines. Its improvement over Outcome SFT suggests that the model benefits from mechanistic supervision when predicting responses in unfamiliar cellular and perturbation contexts. Training-budget controls further support this interpretation: CoT SFT outperforms both sample-matched and token-matched outcome-only continuations (Appendix~\ref{app:budget_controls}), suggesting that additional outcome-only training alone does not reproduce the benefits of the full mechanistic SFT stage.

The genetic-to-chemical fold evaluates chemical responses in HepG2, a context encountered through genetic perturbations during training. On this fold, Outcome SFT reaches 0.482 balanced accuracy, CoT SFT reaches 0.608, and GRPO reaches 0.609, compared with 0.341 for Qwen3-4B and 0.382 for SUMMER. These results support the benefit of mechanistic supervision for this transfer setting, with similar performance for SFT and GRPO. The single-to-combo fold evaluates CRISPRa combinations, none of which are used for training. It jointly tests changes in perturbation composition, intervention modality, and data source. On this fold, Outcome SFT achieves 0.554 balanced accuracy, compared with 0.535 for CoT SFT and 0.537 for GRPO; the overall gains from reasoning supervision therefore do not extend uniformly to this setting.

\begin{table}[t]
\caption{Comparing outcome accuracy and mechanism faithfulness across LLM-based models under the \textit{no path} and \textit{cell-conditioned path (default)} context. All metrics are macro-averaged over the ten test folds. Reason correctness requires nonempty triplets, path connectivity, and edge recall $\geq 0.5$.}
\vspace{-0.02cm}
\label{tab:acc_vs_faith_ctx}
\renewcommand{\arraystretch}{1.2}
\shrinktable{\textwidth}{%
\begin{tabular}{llrrrrrr}
\hline
 &  & \multicolumn{1}{c}{Outcome} & \multicolumn{3}{c}{Reasoning} & \multicolumn{2}{c}{Failure Modes} \\ \cline{3-8} 
\multirow{-2}{*}{\begin{tabular}[c]{@{}l@{}}Pathway\\ Contexts\end{tabular}} & \multirow{-2}{*}{Model} & \multicolumn{1}{l}{Bal. Acc.} & \multicolumn{1}{l}{Edge Recall} & \multicolumn{1}{l}{Path Conn.} & \multicolumn{1}{l}{GO Sim.} & \multicolumn{1}{l}{Answer $\times$ Reason $\checkmark$} & \multicolumn{1}{l}{Answer $\checkmark$ Reason $\times$} \\ \hline
 & SUMMER-4B & \cellcolor[HTML]{EDABAB}0.378 & \cellcolor[HTML]{EA9999}0.097 & \cellcolor[HTML]{EA9999}0.254 & \cellcolor[HTML]{EA9999}0.337 & 0.007 & 0.340 \\
 & Qwen3-4B & \cellcolor[HTML]{EA9999}0.343 & \cellcolor[HTML]{EBA0A0}0.135 & \cellcolor[HTML]{FBEEEE}0.865 & \cellcolor[HTML]{F5D0D0}0.606 & 0.032 & 0.269 \\ \cline{2-8}
 & \PertReasonLM-4B SFT & \cellcolor[HTML]{B3CCF6}0.677 & \cellcolor[HTML]{F4CACA}0.342 & \cellcolor[HTML]{AFC9F6}0.991 & \cellcolor[HTML]{FBEFEF}0.753 & 0.099 & 0.460 \\
\multirow{-4}{*}{No Path} & \multicolumn{1}{r}{\quad + GRPO} & \cellcolor[HTML]{A9C6F5}0.695 & \cellcolor[HTML]{F6D4D4}0.392 & \cellcolor[HTML]{A4C2F4}0.992 & \cellcolor[HTML]{FCF3F3}0.773 & 0.111 & 0.424 \\ \hline
 & SUMMER-4B & \cellcolor[HTML]{EFB5B5}0.397 & \cellcolor[HTML]{CCDDF9}0.818 & \cellcolor[HTML]{FDF6F6}0.919 & \cellcolor[HTML]{ABC7F5}0.930 & 0.475 & 0.038 \\
 & Qwen3-4B & \cellcolor[HTML]{ECA4A4}0.364 & \cellcolor[HTML]{C6D9F9}0.840 & \cellcolor[HTML]{C3D7F8}0.989 & \cellcolor[HTML]{D0E0FA}0.886 & 0.526 & 0.026 \\ \cline{2-8}
 & \PertReasonLM-4B SFT & \cellcolor[HTML]{B1CBF6}0.681 & \cellcolor[HTML]{A4C2F4}0.979 & \cellcolor[HTML]{FEFEFE}0.977 & \cellcolor[HTML]{A4C2F4}0.938 & 0.291 & 0.018 \\
\multirow{-4}{*}{\begin{tabular}[c]{@{}l@{}}Cell-Cond.\\ (default)\end{tabular}} & \multicolumn{1}{r}{\quad + GRPO} & \cellcolor[HTML]{A4C2F4}0.704 & \cellcolor[HTML]{A4C2F4}0.979 & \cellcolor[HTML]{B9D0F7}0.990 & \cellcolor[HTML]{A4C2F4}0.938 & 0.277 & 0.008 \\ \hline
\end{tabular}%
}
\vspace{-0.05cm}
\end{table}

\begin{figure}[t]
  \centering
  \begin{minipage}{0.42\textwidth}
    \centering
    \includegraphics[width=\linewidth]{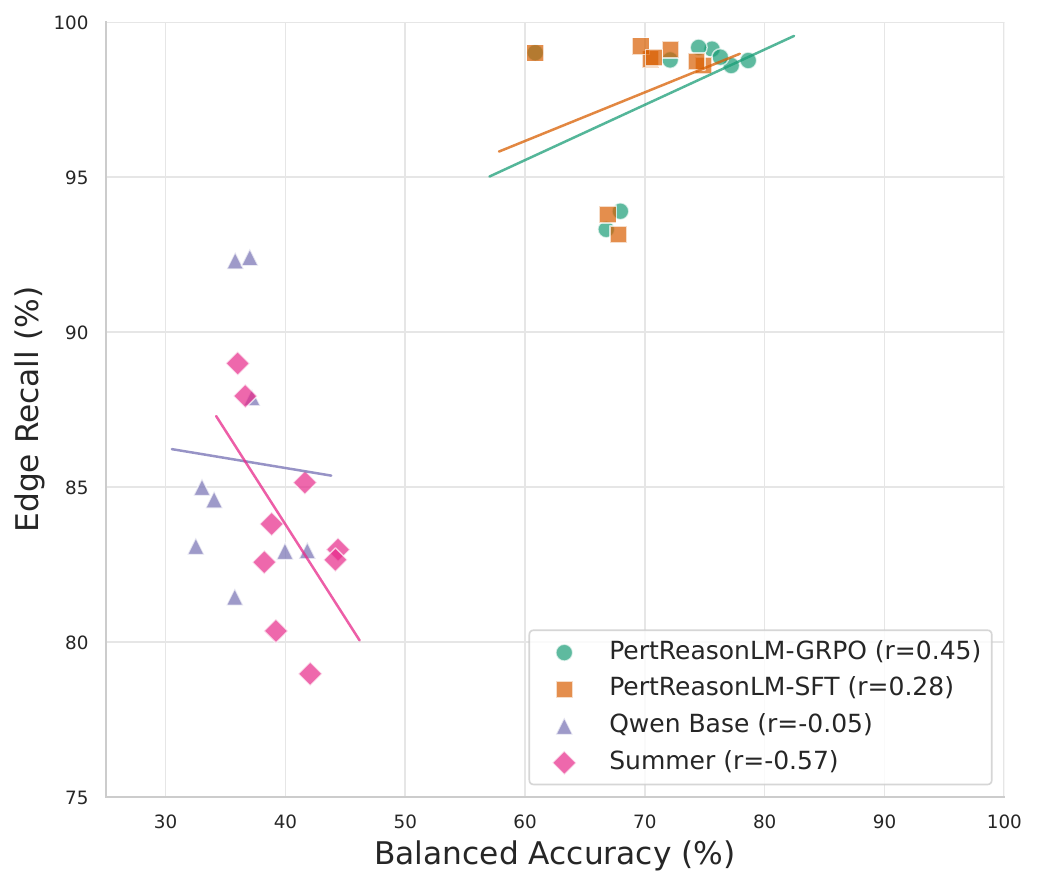}
    \caption{Per-split relationship between balanced accuracy and edge recall under the cell-conditioned path context, excluding single-to-combo.}
    \label{fig:edge_recall_ba_selected}
  \end{minipage}%
  \hfill
  \begin{minipage}{0.54\textwidth}
    \centering
    \includegraphics[width=\linewidth]{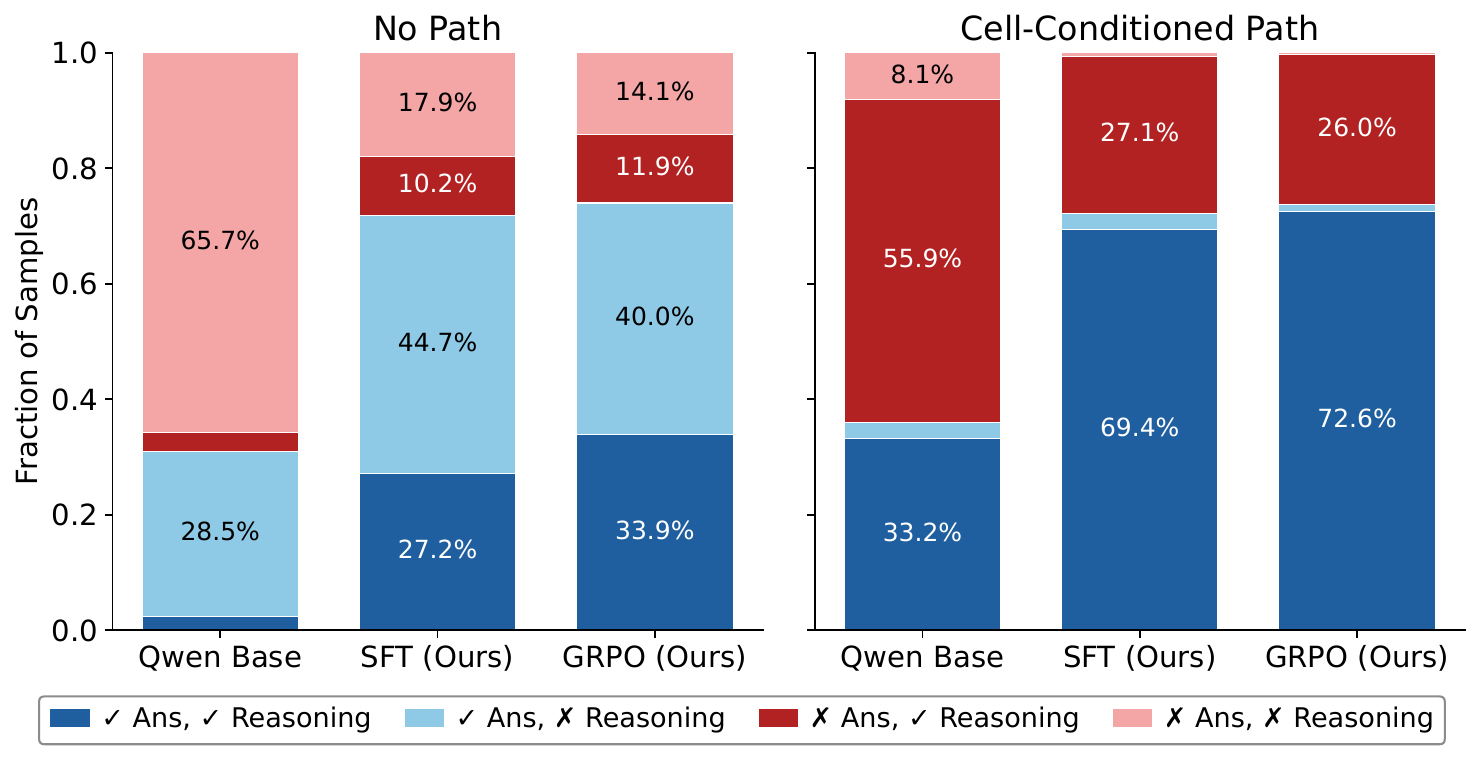}
    \vspace{-0.65cm}
    \caption{Answer--reason taxonomy distribution for Qwen3-4B base, \PertReasonLM-SFT, and \PertReasonLM-GRPO in the scenarios with the no path context and with the cell-conditioned path context. The four colors indicate joint answer/reason correctness, where reason correctness requires triplet existence, path connectivity, and edge recall $\geq 0.5$. Counts are pooled over all ten test folds.}
    \label{fig:error_taxonomy_hidden_noisy}
  \end{minipage}
  \vspace{-0.2cm}
\end{figure}

\vspace{-0.15cm}
\subsection{Benchmarking Reasoning Faithfulness}
\vspace{-0.09cm}

We next investigate whether improved outcome performance is accompanied by faithful reasoning. Table~\ref{tab:acc_vs_faith_ctx} summarizes the mechanistic faithfulness alongside outcome metrics macro-averaged across the ten test folds. We first analyze the cell-conditioned path scenario. Providing pathway contexts generally improves reasoning metrics across all four models, as well as outcome accuracy. While Qwen3-4B and the RAG-based SUMMER successfully extract plausible biological chains, they frequently fail to connect these chains to the correct downstream impact. This disconnect results in high rates of mismatched answers and reasoning.

In contrast, \PertReasonLM improves both outcome accuracy and reasoning faithfulness over the baselines. Supplementary diagnostics further show that its longer traces achieve higher recall and F1 with precision comparable to Qwen3-4B (Appendix~\ref{app:metric_diagnostics}). From \PertReasonLM-SFT to \PertReasonLM-GRPO, accuracy improves while edge recall and GO similarity remain nearly unchanged. The reinforcement learning stage suppresses the ``right answer with wrong reason'' failure mode, reducing it from 1.8\% to 0.8\% (macro-averaged across folds). The advantages of \PertReasonLM also hold in the no path scenario. Without pathway contexts, SUMMER and Qwen3-4B have lower edge recall and GO similarity. Conversely, \PertReasonLM retains higher reasoning connectivity and outcome accuracy even without explicit pathways, although correct answers with wrong reasons remain common.

Figures~\ref{fig:edge_recall_ba_selected} and~\ref{fig:error_taxonomy_hidden_noisy} provide complementary views of mechanism recovery and outcome prediction. Across the nine test folds excluding single-to-combo, edge recall and balanced accuracy are positively associated for \PertReasonLM-SFT and \PertReasonLM-GRPO. Single-to-combo exhibits a different metric profile and involves a distinct intervention modality (CRISPRa versus CRISPRi).
The four-level taxonomy shows that training shifts a large fraction of predictions into the fully correct \textit{Ans $\checkmark$ Reason $\checkmark$} state. 
These results show improved agreement between structured mechanisms and perturbation outcomes. GRPO provides an additional refinement by increasing fully correct predictions and reducing answer--reason disagreement under the cell-conditioned path context.

\vspace{-0.15cm}
\subsection{Ablations by Pathway Contexts, Training Perturbation Modalities, and Model Scales}
\vspace{-0.09cm}

\begin{table}[t]
\caption{Balanced accuracy of \PertReasonLM-SFT on outcome prediction across different pathway contexts. Averages are macro means over all ten folds.}
\vspace{-0.15cm}
\label{tab:by_ctxs_v3}
\setlength{\tabcolsep}{2.5pt}
\shrinktable{\textwidth}{%
\begin{tabular}{lccccccccccc}
\toprule
                                   &                               & \multicolumn{4}{c}{ID Cell}                                                                                                   & \multicolumn{4}{c}{OOD Cell} & & \\ \cmidrule(lr){3-10}
\multirow{-2}{*}{Pathway Contexts} & \multirow{-2}{*}{Average}     & ID Chem.                      & ID Gen.                       & OOD Chem.                     & OOD Gen.                      & ID Chem.                      & ID Gen.                       & OOD Chem.                     & OOD Gen. & \multirow{-2}{*}{\begin{tabular}[c]{@{}c@{}}Gen-to\\ Chem\end{tabular}} & \multirow{-2}{*}{\begin{tabular}[c]{@{}c@{}}Single-to\\ Combo\end{tabular}} \\ \midrule
No Path & \cellcolor[HTML]{FEF8F8}0.677 & \cellcolor[HTML]{D9E6FA}0.752 & \cellcolor[HTML]{FCFDFF}0.692 & \cellcolor[HTML]{FDF6F6}0.674 & \cellcolor[HTML]{E5EDFC}0.732 & \cellcolor[HTML]{EFF5FD}0.714 & \cellcolor[HTML]{F2F6FD}0.710 & \cellcolor[HTML]{FDF4F4}0.671 & \cellcolor[HTML]{F2F6FD}0.710 & \cellcolor[HTML]{F1BBBB}0.584 & \cellcolor[HTML]{EA9999}0.533 \\
Uncond. Paths & \cellcolor[HTML]{FCF1F1}0.666 & \cellcolor[HTML]{D1E0F9}0.766 & \cellcolor[HTML]{FAFCFE}0.696 & \cellcolor[HTML]{FCFDFF}0.693 & \cellcolor[HTML]{F8DCDC}0.634 & \cellcolor[HTML]{FAFCFE}0.696 & \cellcolor[HTML]{F9FBFE}0.698 & \cellcolor[HTML]{FBECEC}0.659 & \cellcolor[HTML]{FFFFFF}0.688 & \cellcolor[HTML]{F2BFBF}0.591 & \cellcolor[HTML]{EA9B9B}0.536 \\
Cell-Cond. Path (Default) & \cellcolor[HTML]{FEFBFB}0.681 & \cellcolor[HTML]{DBE7FB}0.749 & \cellcolor[HTML]{EBF1FD}0.722 & \cellcolor[HTML]{FEF9F9}0.678 & \cellcolor[HTML]{F3F7FE}0.708 & \cellcolor[HTML]{F5F8FE}0.705 & \cellcolor[HTML]{F9FBFE}0.697 & \cellcolor[HTML]{FCF3F3}0.669 & \cellcolor[HTML]{DEE9FB}0.743 & \cellcolor[HTML]{F4CBCB}0.608 & \cellcolor[HTML]{EA9A9A}0.535 \\
Reference Path (Oracle) & \cellcolor[HTML]{C1D6F8}0.792 & \cellcolor[HTML]{A4C2F4}0.842 & \cellcolor[HTML]{C1D5F7}0.793 & \cellcolor[HTML]{B2CBF6}0.818 & \cellcolor[HTML]{C4D7F8}0.788 & \cellcolor[HTML]{C0D5F7}0.795 & \cellcolor[HTML]{B3CCF6}0.817 & \cellcolor[HTML]{B1CBF6}0.820 & \cellcolor[HTML]{BDD3F7}0.799 & \cellcolor[HTML]{D7E4FA}0.756 & \cellcolor[HTML]{FCFDFF}0.693 \\ \bottomrule
\end{tabular}%
}
\vspace{-0.05cm}
\end{table}

\begin{table}[t]
\caption{Balanced accuracy of \PertReasonLM-SFT under the cell-conditioned path context. The shared default (4B with all supervision) is the reference for (a) training perturbation modalities and (b) model scales. Averages are macro means over the ten folds.}
\vspace{-0.15cm}
\label{tab:ablation}
\setlength{\tabcolsep}{3.5pt}
\shrinktable{\textwidth}{%
\begin{tabular}{lccccccccccc}
\toprule
 &  & \multicolumn{4}{c}{ID Cell} & \multicolumn{4}{c}{OOD Cell} & & \\ \cmidrule(lr){3-10}
\multirow{-2}{*}{Configuration} & \multirow{-2}{*}{Average} & ID Chem. & ID Gen. & OOD Chem. & OOD Gen. & ID Chem. & ID Gen. & OOD Chem. & OOD Gen. & \multirow{-2}{*}{\begin{tabular}[c]{@{}c@{}}Gen-to\\ Chem\end{tabular}} & \multirow{-2}{*}{\begin{tabular}[c]{@{}c@{}}Single-to\\ Combo\end{tabular}} \\ \midrule
Default (All, 4B) & \cellcolor[HTML]{E1EBFB}0.681 & \cellcolor[HTML]{B6CEF6}0.749 & \cellcolor[HTML]{C7DAF8}0.722 & \cellcolor[HTML]{E3ECFC}0.678 & \cellcolor[HTML]{D0DFF9}0.708 & \cellcolor[HTML]{D2E1FA}0.705 & \cellcolor[HTML]{D7E4FA}0.697 & \cellcolor[HTML]{E8F0FC}0.669 & \cellcolor[HTML]{BAD1F7}0.743 & \cellcolor[HTML]{FDF6F6}0.608 & \cellcolor[HTML]{F8DDDD}0.535 \\
\midrule
\multicolumn{12}{l}{(a) Training perturbation modalities} \\
\addlinespace[2pt]
Chemical & \cellcolor[HTML]{FDF5F5}0.605 & \cellcolor[HTML]{D7E4FA}0.697 & \cellcolor[HTML]{FAE6E6}0.562 & \cellcolor[HTML]{E2ECFC}0.679 & \cellcolor[HTML]{FBEAEA}0.572 & \cellcolor[HTML]{F4F7FE}0.651 & \cellcolor[HTML]{FCF1F1}0.594 & \cellcolor[HTML]{F7FAFE}0.646 & \cellcolor[HTML]{FEF9F9}0.617 & \cellcolor[HTML]{FCEFEF}0.588 & \cellcolor[HTML]{F2BEBE}0.447 \\
Genetic & \cellcolor[HTML]{FCF1F1}0.592 & \cellcolor[HTML]{F8DCDC}0.534 & \cellcolor[HTML]{CFDFF9}0.710 & \cellcolor[HTML]{FBEDED}0.582 & \cellcolor[HTML]{D6E4FA}0.698 & \cellcolor[HTML]{F8DBDB}0.530 & \cellcolor[HTML]{F2F7FD}0.653 & \cellcolor[HTML]{F9E2E2}0.550 & \cellcolor[HTML]{E7EFFC}0.671 & \cellcolor[HTML]{F4C9C9}0.480 & \cellcolor[HTML]{F6D6D6}0.515 \\
\midrule
\multicolumn{12}{l}{(b) Model scales} \\
\addlinespace[2pt]
1.7B & \cellcolor[HTML]{FDF7F7}0.611 & \cellcolor[HTML]{E4EDFC}0.676 & \cellcolor[HTML]{FDF3F3}0.599 & \cellcolor[HTML]{ECF2FD}0.663 & \cellcolor[HTML]{FDF3F3}0.599 & \cellcolor[HTML]{FFFFFF}0.633 & \cellcolor[HTML]{FBEAEA}0.574 & \cellcolor[HTML]{FDF4F4}0.601 & \cellcolor[HTML]{FCF1F1}0.593 & \cellcolor[HTML]{FDF6F6}0.606 & \cellcolor[HTML]{FAE7E7}0.564 \\
8B & \cellcolor[HTML]{EFF4FD}0.658 & \cellcolor[HTML]{A4C2F4}0.778 & \cellcolor[HTML]{DFEAFB}0.684 & \cellcolor[HTML]{CDDEF9}0.712 & \cellcolor[HTML]{BCD2F7}0.739 & \cellcolor[HTML]{C6D9F8}0.724 & \cellcolor[HTML]{EEF4FD}0.660 & \cellcolor[HTML]{FEFAFA}0.620 & \cellcolor[HTML]{BED4F7}0.736 & \cellcolor[HTML]{FBEDED}0.583 & \cellcolor[HTML]{EA9999}0.342 \\
\bottomrule
\end{tabular}%
}
\vspace{-0.30cm}
\end{table}

\paragraph{Pathway Contexts}

Table~\ref{tab:by_ctxs_v3} shows that cell-conditioned pathway contexts improve average balanced accuracy over unconditional paths, which perform worse than no paths, suggesting that cellular conditioning helps limit irrelevant ``evidence''. The strong performance even in the no path scenario indicates that the model already recalls perturbation-response regularities learned during training, while the cell-conditioned path complements this internal knowledge with cell-specific mechanistic context. For OOD genetic perturbations in OOD cells, the cell-conditioned path scenario achieves 0.743 balanced accuracy, outperforming both scenarios of no path (0.710) and unconditional paths (0.688). Reference paths further raise performance, suggesting room for improvement in pathway context quality, although this oracle setting requires outcome knowledge.

\vspace{-0.15cm}
\paragraph{Training Perturbation Modalities}

Table~\ref{tab:ablation}(a) compares chemical-only and genetic-only supervision against the shared default (joint supervision), with the 4B model fixed. Joint chemical and genetic supervision achieves the highest average balanced accuracy across the ten folds (0.681), compared with chemical-only (0.605) and genetic-only (0.592) supervision, and leads on nine of the ten individual folds. These results suggest that mixed supervision helps \PertReasonLM learn regulatory patterns that generalize across perturbation modalities and evaluation settings.

\vspace{-0.15cm}
\paragraph{Model Scales}

Table~\ref{tab:ablation}(b) compares the 1.7B and 8B models against the shared 4B default. Average balanced accuracy increases from 0.611 at 1.7B to 0.681 at 4B, then decreases to 0.658 at 8B. The 8B model leads on three ID-cell folds and ID chemical perturbations in OOD cells, while the 1.7B model performs best on single-to-combo. The 4B model achieves the highest average balanced accuracy, while increasing model scale does not yield uniform gains across generalization settings.

\vspace{-0.15cm}
\subsection{Biological Support for Reference Mechanisms and Model-Generated Traces}
\vspace{-0.09cm}

\paragraph{Biological Support for Reference Mechanisms}

A biology expert with single-cell/pathology expertise audited 160 frozen \PertReasonQA references against supporting and contradictory literature, judging 58.1\% supported, 38.8\% uncertain, and 3.1\% unsupported. Evidence for all required interactions and the exact cellular context was more limited. These findings support using the references as evidence-grounded mechanistic proxies with explicit uncertainty, interpreted in light of the available pathway and contextual evidence. They provide biologically informed routes within a broader set of possible mechanisms. The audit criteria and full results are detailed in Appendix~\ref{app:reference_audit}.

\vspace{-0.15cm}
\paragraph{Biological Support for Model-Generated Traces}

To assess the biological support of generated explanations, we conducted a rule-based review covering 960 judgments (five model-generated traces and one corrupted control for each of 160 shared queries). Explanations from both \PertReasonLM-SFT and GRPO received substantial biological support. Their similar support judgments complement the earlier finding that GRPO improves outcome accuracy and answer--reason agreement while largely preserving reference-edge recovery. Structurally coherent corrupted controls received little biological support, showing the value of checking biological evidence alongside chain structure. Appendix~\ref{app:trace_audit} describes the review design and compares biological support.

\vspace{-0.15cm}
\paragraph{Interpreting Automatic Metrics}

Explanations recovering more reference interactions and matching regulatory directions tended to receive stronger biological support. GO similarity was weakly negatively correlated with biological support judgments and is used only to describe broad functional relatedness. Together, these findings support interpreting automatic scores alongside biological evidence; valid alternative routes may receive lower reference-overlap scores. For the audit and metric analyses, see Appendix~\ref{app:trace_audit}.

\vspace{-0.15cm}
\subsection{Discussion and Limitations}
\vspace{-0.09cm}

\paragraph{Beyond Outcome-Centric Perturbation Evaluation}

\PertReasonQA highlights that perturbation models should be evaluated not only by whether they predict the correct expression outcome, but also by whether the prediction is supported by a biologically plausible mechanism. By pairing perturbation outcomes with cell-state-conditioned regulatory pathways, \PertReasonQA complements existing perturbation prediction suites by assessing reference consistency and exposing answer--reason disagreement under realistic shifts.

\vspace{-0.15cm}
\paragraph{Scope and Coverage of Reasoning Evaluation}

Our reasoning evaluation covers only queries with a KG reference path that agrees with the observed outcome, which may favor well-studied genes and pathways. SFT also improves outcome prediction when retrieved paths do not match the outcome or no path is retrieved. These findings suggest that the observed outcome gains extend beyond queries with reference paths within the sampled evaluations; broader query coverage and reasoning assessment without references remain directions for future evaluation. See Appendix~\ref{app:coverage_evaluation} for the supporting evaluation.

\vspace{-0.15cm}
\paragraph{Limitations and Future Directions for Mechanistic Supervision}

\PertReasonQA converts structured causal paths into natural-language texts using Qwen3-4B, which may introduce stylistic biases or favor models with similar linguistic priors. Training settings established for the 4B model and reused across scales may also have contributed to its strong performance in the scale ablation. Future evaluations should incorporate expert-written rationales, alternative verbalization methods, evidence provenance, and multiple valid mechanistic explanations per perturbation-effect pair.

\vspace{-0.17cm}
\section{Conclusion}
\vspace{-0.15cm}

We introduced \PertReasonQA, a knowledge-grounded benchmark for evaluating whether models can reason faithfully about perturbation effects under realistic distribution shifts, including genetic-to-chemical transfer and single-to-combo generalization. By pairing perturbation outcomes with cell-conditioned regulatory pathways, \PertReasonQA moves evaluation beyond label accuracy and exposes failure modes that existing perturbation benchmarks often miss. We further presented \PertReasonLM as a reference probe that aligns outcome prediction with context-specific mechanistic explanations, showing that reasoning-centric training can reduce these gaps. Together, these results suggest that progress in virtual cell modeling should be measured not only by predictive performance, but also by the ability to produce biologically plausible and generalizable mechanisms.

\clearpage

\begin{ack}
This research was in part supported by the National Institute of General Medical Sciences of the National Institutes of Health under award number R35GM124952. 
Portions of this research were conducted with advanced computing resources provided by Texas A\&M High Performance Research Computing.
\end{ack}

\bibliographystyle{plainnat}
{
\small
\bibliography{ref}
}


\clearpage
\appendix



\etocdepthtag.toc{appendix}

\begingroup
\setlength{\parskip}{0pt}
\renewcommand{\contentsname}{Appendix Contents}
\etocsettagdepth{main}{none}
\etocsettagdepth{appendix}{subsection}
\etocsettagdepth{checklist}{none}
\tableofcontents
\endgroup
\clearpage

\raggedbottom

\section{Data Details}\label{app:data}
This section details the dataset statistics, evaluation splits, and construction of cell-conditioned reference reasoning in \PertReasonQA.

\renewcommand{\arraystretch}{1.1}
\begin{table}[t]
\caption{Full summary statistics of \PertReasonQA across training, validation, and ten test folds. Perturbation counts are unique within each sample type and column. Outcome-only samples are used only for training and validation.}
\vspace{-0.09cm}
\label{tab:data_stats_full}
\setlength{\tabcolsep}{2pt}
\shrinktable{\textwidth}{%
\begin{tabular}{llrrrrrrrrrrrrrr}
\toprule
\multicolumn{2}{l}{} & \multicolumn{2}{c}{Train ID Cell} & \multicolumn{2}{c}{Valid ID Cell} & \multicolumn{4}{c}{Test ID Cell} & \multicolumn{4}{c}{Test OOD Cell} & \multicolumn{1}{c}{\multirow{2}{*}{\begin{tabular}[c]{@{}c@{}}Gen-to\\ Chem\end{tabular}}} & \multicolumn{1}{c}{\multirow{2}{*}{\begin{tabular}[c]{@{}c@{}}Single-to\\ Combo\end{tabular}}} \\ \cmidrule(lr){3-14}
\multicolumn{2}{l}{} & \multicolumn{1}{c}{Chem.} & \multicolumn{1}{c}{Gen.} & \multicolumn{1}{c}{Chem.} & \multicolumn{1}{c}{Gen.} & \multicolumn{1}{c}{ID Chem.} & \multicolumn{1}{c}{ID Gen.} & \multicolumn{1}{c}{OOD Chem.} & \multicolumn{1}{c}{OOD Gen.} & \multicolumn{1}{c}{ID Chem.} & \multicolumn{1}{c}{ID Gen.} & \multicolumn{1}{c}{OOD Chem.} & \multicolumn{1}{c}{OOD Gen.} & & \\ \midrule
\multicolumn{2}{l}{Cellular Contexts} & 5 & 3 & 5 & 3 & 5 & 3 & 5 & 3 & 2 & 1 & 2 & 1 & 1 & 1 \\ \midrule
\multirow{4}{*}{\begin{tabular}[c]{@{}l@{}}Outcome-only\\ Samples\end{tabular}}
  & Perturbations & 135 & 1172 & 17 & 137 & -- & -- & -- & -- & -- & -- & -- & -- & -- & -- \\
  & Unchanged & 62730 & 212216 & 7583 & 24042 & -- & -- & -- & -- & -- & -- & -- & -- & -- & -- \\
  & Up & 32304 & 177586 & 2600 & 22951 & -- & -- & -- & -- & -- & -- & -- & -- & -- & -- \\
  & Down & 53412 & 281349 & 6653 & 31775 & -- & -- & -- & -- & -- & -- & -- & -- & -- & -- \\ \midrule
\multirow{4}{*}{\begin{tabular}[c]{@{}l@{}}Reasoning\\ Samples\end{tabular}}
  & Perturbations & 127 & 636 & 17 & 69 & 14 & 67 & 21 & 15 & 14 & 62 & 19 & 12 & 4 & 5 \\
  & Unchanged & 19111 & 20123 & 2678 & 2347 & 2316 & 1275 & 1913 & 62 & 809 & 192 & 815 & 24 & 1788 & 543 \\
  & Up & 17730 & 13000 & 1327 & 2357 & 1579 & 1265 & 628 & 133 & 803 & 396 & 496 & 64 & 388 & 147 \\
  & Down & 29122 & 26817 & 3149 & 4085 & 3252 & 1939 & 3203 & 464 & 1406 & 692 & 1935 & 273 & 467 & 210 \\ \midrule
\multicolumn{2}{l}{Avg. Pathway Length} & 5.98 & 4.55 & 6.11 & 4.22 & 6.25 & 4.68 & 5.29 & 3.95 & 6.05 & 4.89 & 5.07 & 4.35 & 6.05 & 6.01 \\ \bottomrule
\end{tabular}%
}
\end{table}

\subsection{Dataset Statistics}\label{app:data_stats}

Table~\ref{tab:data_stats_full} reports the full dataset statistics, including outcome-only samples used exclusively for training and validation and reasoning samples across all splits.

\subsection{Splits}\label{app:splits}

To evaluate generalizability, we construct ten test folds. Eight combine ID/OOD cellular contexts with ID/OOD perturbation clusters separately for chemical and genetic modalities, allowing us to evaluate extrapolation to novel cells, novel perturbations, and their joint shift. We define OOD cellular contexts independently of the perturbation split by holding out complete cellular contexts: T cells and A549 for chemical perturbations, and RPE1 for genetic perturbations.

The genetic-to-chemical fold uses chemical perturbations in HepG2 from DeMeo et al.~\citep{demeo2025active}. HepG2 is observed during genetic training but held out from chemical training, testing transfer of cellular context across perturbation modalities. Chemical training examples from other cellular contexts remain available in the mixed-modality training set. 

The single-to-combo fold uses K562 CRISPRa combinations from Norman et al.~\citep{norman2019exploring} exclusively for testing. No Norman samples, whether single or combinatorial, enter training or validation; training uses single perturbations from the other sources. This setting jointly changes the number of targeted genes, the genetic intervention modality (CRISPRi to CRISPRa), and the data source, so its results cannot isolate combinatorial generalization alone.

To rigorously evaluate generalization, we implement a functional splitting strategy for both genetic and chemical perturbations. By partitioning perturbations based on biological function rather than random assignment, we aim to probe mechanistic extrapolation while limiting opportunities to memorize overlapping pathways. We construct a heterogeneous directed biological graph from OmniPath~\citep{turei2016omnipath} and augment it with Gene Ontology (GO) gene sets from MSigDB (v2023.2)~\citep{liberzon2015molecular}. We then compute Personalized PageRank (PPR) scores across the gene interaction network and aggregate them onto connected GO term nodes, yielding a gene-to-GO continuous vector encoding each gene's functional and structural context.

Based on cosine similarity of these functional vectors, we apply KMeans clustering ($k=30$) to group genes sharing similar biological roles, then randomly partition entire clusters into ID and OOD sets. A genetic perturbation inherits the assignment of its perturbed gene. Chemical perturbations are aligned to this functional split by mapping drugs to their target genes via INDRA CoGEx~\citep{indra}; a drug is assigned to OOD if more than 10\% of its target genes fall into OOD gene clusters, and to ID otherwise (at most 10\%). Thus, evaluation with OOD perturbations probes mechanistic extrapolation under a functional distribution shift, while evaluation in OOD cellular contexts tests generalization to unseen basal cellular contexts. This formulation guarantees that both genetic and chemical OOD splits systematically evaluate the model’s capacity under functional distribution shift.

\subsection{Knowledge-Grounded Reasoning Details}\label{app:reasoning}

We detail the construction of cell-conditioned reference reasoning, from basal activity thresholding and pathway extraction to consistency filtering and text synthesis.

\subsubsection{Cell-Specific Basal Activity Thresholding}\label{app:threshold}

A fundamental step in modeling cell-specific mechanisms is interpreting the continuous spectrum of basal gene expression. We use log-transformed single-cell expression within each cell context and compare perturbed cells with matched controls. Raw transcriptomic data frequently suffer from technical noise and systemic batch effects across different experimental runs, making fixed global thresholds highly unreliable. To robustly categorize expression levels across diverse contexts, we establish adaptive, cell-specific thresholds. Operationally, we identify two critical transition points in the data density. The low threshold ($\tau_{\mathrm{low}}$) marks a data-adaptive lower boundary for basal activity. Conversely, the high threshold ($\tau_{\mathrm{high}}$) marks the upper transition of typical active expression. This dynamic thresholding mitigates batch effects and provides a normalized context for the \PertReasonQA task.

Threshold estimation follows a split-aware supervised calibration protocol. We first derive cell-context-specific calibration targets from the changed and unchanged groups defined by the differential-expression procedure on the training set only, and fit separate gradient-boosted regression tree models for $\tau_{\mathrm{low}}$ and $\tau_{\mathrm{high}}$ using the eight training contexts. Each model takes 101 basal-expression percentiles (0th through 100th) as input and uses 500 boosting stages, and a learning rate of 0.05. Calibration is specific to each cell context within its data source. For evaluation, thresholds are fixed and reused for contexts observed during training, while those for unseen contexts are predicted solely from their basal expression quantiles (Table~\ref{tab:basal_thresholds}). Only these fixed or predicted thresholds are used for graph construction and prompt generation; neither test changed/unchanged labels nor perturbation-induced expression outcomes enter calibration or prediction. Accordingly, the evaluation assumes access to a basal control profile for each test cell context.

\begin{table}[t]
\centering
\caption{Basal activity thresholds on the log-transformed expression scale, by fold group and cell context. ID/OOD cell groups use the same thresholds for ID and OOD perturbations. For cell contexts seen during training, thresholds remain fixed at their training-calibrated values. For unseen cell contexts, thresholds are predicted from basal control expression alone.}
\label{tab:basal_thresholds}
\small
\begin{tabular}{lllrr}
\toprule
Fold group & Cell context & Threshold use & $\tau_{\mathrm{low}}$ & $\tau_{\mathrm{high}}$ \\
\midrule
\multirow{5}{*}{ID Cell / Chem.} & B cells & Fixed & 0.216418 & 1.616057 \\
 & Myeloid cells & Fixed & 0.164922 & 1.453653 \\
 & NK cells & Fixed & 0.255494 & 3.665738 \\
 & K562 & Fixed & 0.075223 & 1.279859 \\
 & MCF7 & Fixed & 0.050304 & 0.672142 \\
\midrule
\multirow{2}{*}{OOD Cell / Chem.} & T cells & Predicted & 0.216419 & 1.564164 \\
 & A549 & Predicted & 0.071699 & 1.201166 \\
\midrule
\multirow{3}{*}{ID Cell / Gen.} & HepG2 & Fixed & 0.322951 & 1.554728 \\
 & Jurkat & Fixed & 0.424008 & 1.638669 \\
 & K562 & Fixed & 0.302047 & 1.624157 \\
\midrule
OOD Cell / Gen. & RPE1 & Predicted & 0.319688 & 1.620234 \\
Genetic-to-chemical & HepG2 & Predicted & 0.157658 & 1.454501 \\
Single-to-combo & K562 & Predicted & 0.183972 & 1.464625 \\
\bottomrule
\end{tabular}
\end{table}

To determine the low-threshold calibration target for each training cell context $c$, we employ a maximum separability criterion using a two-sample Kolmogorov-Smirnov (KS) statistic. We group basal expression values according to whether genes are labeled unchanged or changed by the training differential-expression procedure. Denoting the resulting empirical cumulative distribution functions (ECDFs) by $\hat{F}_{\mathrm{unchanged},c}^{\mathrm{train}}(x)$ and $\hat{F}_{\mathrm{changed},c}^{\mathrm{train}}(x)$, the target is defined as:
$$\tau_{\mathrm{low},c}^{\mathrm{train}} = \underset{x}{\operatorname{arg\,max}} \left( \hat{F}_{\mathrm{unchanged},c}^{\mathrm{train}}(x) - \hat{F}_{\mathrm{changed},c}^{\mathrm{train}}(x) \right)$$
This objective identifies the expression value at which the two empirical training label groups are maximally separated.

To capture the upper bounds of typical expression dynamics, we define the high threshold ($\tau_{\mathrm{high}}$), which represents the biological saturation point of gene expression. For genes exhibiting active expression ($X_{\mathrm{act}} > \tau_{\mathrm{low}}$), we apply a normalization mapping $\phi: X_{\mathrm{act}} \to [0, 1]$:
$$z_i = \phi(x_i) = \frac{x_i - \min(X_{\mathrm{act}})}{\max(X_{\mathrm{act}}) - \min(X_{\mathrm{act}})}$$
Let $\hat{F}_{\mathrm{act}}(z)$ denote the ECDF of these normalized values. We compare this empirical distribution against the cumulative distribution function of a standard Uniform distribution, $F_{U}(z) = z$, which acts as a constant density baseline. The high threshold is determined by calculating the one-sample Kolmogorov-Smirnov (KS) statistic against this linear baseline:
$$z^* = \underset{z \in [0, 1]}{\operatorname{arg\,max}} \left| \hat{F}_{\mathrm{act}}(z) - z \right|,\quad \tau_{\mathrm{high}} = \phi^{-1}(z^*)$$
In the context of heavy-tailed biological data, this geometric knee point effectively localizes the structural transition from the main body of active expression to a state of saturation.

\subsubsection{Cell-State-Conditioned Subgraph Extraction}\label{app:subgraph}

General KGs assemble regulatory interactions across diverse cellular states, providing a broad signaling landscape that requires contextualization for a given transcriptomic state. We dynamically refine this static graph into a context-specific subgraph by modulating edge weights according to basal expression.

Let $\gG = (\sV, \sA)$ be the global knowledge graph derived from OmniPath, where each node $v \in \sV$ denotes a gene, and each directed edge $(u,v) \in \sA$ denotes a regulatory relation. For a specific cell state $c$, we retrieve its basal gene expression profile $\mathbf{b}^{(c)} \in \mathbb{R}^{|\sV|}$. 

Using the established low threshold $\tau_{\mathrm{low}}$, we implement an asymmetric basal activity gating mechanism that modulates edge traversability based on gene expression context. For each edge $(u, v)$, we define a structural base weight $w^{\mathrm{base}}_{uv}$, which is then modulated using sigmoid-based gating functions with asymmetric treatment of source and target nodes:
$$\gamma^{\mathrm{src}}_u = \max\Big(\sigma\big(\beta (\mathbf{b}^{(c)}_u - \tau_{\mathrm{low}})\big), \delta_{\mathrm{src}}\Big)$$
$$\gamma^{\mathrm{tgt}}_v = \delta_{\mathrm{tgt}} + (1 - \delta_{\mathrm{tgt}}) \cdot \sigma\big(\beta (\mathbf{b}^{(c)}_v - \tau_{\mathrm{low}})\big)$$
where $\sigma(\cdot)$ denotes the logistic sigmoid function, $\beta$ is a steepness parameter controlling the gate's sensitivity, and $\delta_{\mathrm{src}}$ and $\delta_{\mathrm{tgt}}$ are floor constants that maintain minimum traversability. The final context-conditioned edge weight is:
$$w^{(c)}_{uv} = w^{\mathrm{base}}_{uv} \cdot \gamma^{\mathrm{src}}_u \cdot \gamma^{\mathrm{tgt}}_v + \epsilon_{w}$$
where $\epsilon_{w}$ is a small constant to prevent weights from becoming exactly zero. The source floor $\delta_{\mathrm{src}}$ preserves traversal through weakly expressed genes, while the target floor $\delta_{\mathrm{tgt}}$ preserves reachability of basally silent genes that may be induced after perturbation. Low basal expression therefore down-weights interactions without deleting their nodes or edges.

\subsubsection{Diversity-Aware Causal Path-finding}\label{app:pathfinding}

To construct robust reasoning chains from perturbation sources to differential expression targets, we extended the TieDIE framework, which was originally designed to connect genomic events to transcriptional states through network diffusion. Our approach extends TieDIE with three key modifications: (1) cell-state-conditioned edge weights, (2) explicit path extraction rather than subnetwork identification, and (3) iterative penalty-based diversification.

Given the weighted graph $\gG^{(c)} = (\sV, \sA, \{w^{(c)}_{uv}\})$, we follow influence propagation by estimating a perturbation-centered flow score over nodes using Personalized PageRank (PPR). Let $\mathcal{S}(p)$ denote the set of start nodes associated with perturbation $p$:
$$\mathcal{S}(p) = \begin{cases} \{t : t \text{ is a known target of chemical } p\}, & \text{if } p \text{ is chemical},\\ \{p\}, & \text{if } p \text{ is a single-gene perturbation},\\ \{p_1,\dots,p_m\}, & \text{if } p = p_1 + \dots + p_m \text{ is a gene combination}. \end{cases}$$
We define a personalization vector $\mathbf{s}^{(p)} \in \mathbb{R}^{|\sV|}$ by assigning equal mass to valid start nodes:
$$s^{(p)}_v = \begin{cases} \frac{1}{|\mathcal{S}(p)|}, & \text{if } v \in \mathcal{S}(p),\\ 0, & \text{otherwise}. \end{cases}$$
Let $\mathbf{P}^{(c)}$ be the row-normalized transition matrix induced by the edge weights $w^{(c)}_{uv}$. The node flow score $\mathbf{p}^{(c,p)}$ is computed as the fixed point of:
$$\mathbf{p}^{(c,p)} = (1-\alpha)\mathbf{s}^{(p)} + \alpha (\mathbf{P}^{(c)})^\top \mathbf{p}^{(c,p)}$$
where $\alpha$ is the damping factor representing the probability of continuing the random walk. The resulting scores define a flow field quantifying each node's mechanistic relevance to the perturbation.

Next, we extract explicit causal paths from $\mathcal{S}(p)$ to the effect gene $g_{e}$ using Dijkstra search on a PPR-informed cost function. For each traversable edge $(u,v)$, the search cost is defined as:
$$\mathrm{cost}^{(c,p)}_{uv} = \frac{1}{w^{(c)}_{uv}\,\mathbf{p}^{(c,p)}_u\,\mathbf{p}^{(c,p)}_v + \epsilon_{c}}$$
where $\epsilon_{c}$ is a small cost constant ensuring numerical stability. This cost formulation ensures that paths traverse high-flow regions identified by network diffusion while respecting cell-state structurally plausible edges. The extracted shortest path $\mathcal{P}^\star$ is the one minimizing cumulative resistance $\mathcal{L}(\mathcal{P}) = \sum \mathrm{cost}^{(c,p)}_{v_i v_{i+1}}$, subject to a maximum hop-depth constraint $D_{\max}$. 

To capture the multifaceted nature of perturbation mechanisms, such as compensatory signaling axes, we employ iterative penalty-based diversification. After extracting a path, we apply multiplicative penalties to its visited edges in a temporary graph copy, dividing their weight by a penalty factor $\lambda$ to encourage the exploration of alternative mechanistic routes. Duplicate paths and paths that are complete supersets of other selected paths are pruned. This procedure yields a diverse ensemble of high-confidence causal paths, organized as a perturbation-anchored bag of paths tailored to the queried response gene. Together, basal-state conditioning and perturbation-centered diffusion ground the extracted paths in biologically plausible, cell-specific signaling events. Dijkstra thus optimizes paths on a cell- and perturbation-conditioned weighted graph rather than minimizing unweighted hop count alone.

\subsubsection{Reference Paths via Evidence Augmentation and Filtering}\label{app:pathfiltering}

The primary objective of this stage is to filter the extracted causal paths to eliminate mechanistically invalid reasoning trajectories. However, the paths extracted from OmniPath merely establish structural connectivity and frequently contain edges where the regulatory direction is unknown or functionally ambiguous. To enable the rigorous causal inference required for logical filtering, we must first maximize directional certainty. We achieve this by cross-referencing these unspecified edges with directional literature evidence from INDRA CoGEx. By adopting the majority consensus from the literature, we resolve the ambiguous relations into explicit activating ($+1$) or inhibitory ($-1$) interactions. This evidence augmentation transforms the candidate subgraph into a functionally signed causal network, laying the necessary groundwork for subsequent validation.

This reference-path generation is retrospective benchmark construction, rather than KG-only prediction: observed differential expression labels are used at this stage only to determine which candidate paths are retained as references. Specifically, we apply deterministic filtering criteria based on the observed differential expression labels. For samples exhibiting significant differential expression (labels \texttt{up} or \texttt{down}), a biologically valid path must possess a cumulative polarity that matches the observed outcome. We formalize this using a Sign-Reachability criterion (Algorithm~\ref{alg:reachability}). Biologically, a perturbation initiates a specific signal cascade. As this signal propagates through the network, inhibitory edges invert the polarity. A path is retained only if the terminal signal polarity logically aligns with the observed gene expression shift. Paths that accumulate excessive ambiguity or fundamentally contradict the observed differential state are pruned.

Explaining an \texttt{unchanged} transcriptomic response presents a unique biological challenge, as a lack of differential expression does not necessarily imply an absence of causal interaction. It frequently stems from biological boundary conditions, such as ceiling or floor effects in gene transcription. To model this phenomenon, we introduce a Sign-Consistency framework (Algorithm~\ref{alg:consistency}) that directly integrates the propagated causal signal with the basal transcriptomic state. Utilizing the hysteresis thresholds $\tau_{\mathrm{low}}$ and $\tau_{\mathrm{high}}$ defined in previous sections, we discretize the basal expression of the target gene into distinct functional states: silent, active, or saturated. The algorithm then evaluates whether the incoming perturbation signal can physically alter this state. Crucially, if an activating signal reaches a gene already at its saturation point ($\tau_{\mathrm{high}}$), or if an inhibitory signal reaches a basally silent gene ($\tau_{\mathrm{low}}$), the mechanistic outcome is correctly classified as \texttt{unchanged}. This basal context integration prevents the false rejection of valid causal pathways that are simply masked by physiological constraints.

Directionality is the final consistency check after cell-conditioned path extraction. Graph weights reflect pathway plausibility, not DE magnitude, and statistical significance is distinct from effect size. Given the categorical outcomes and the KG's lack of calibrated interaction strengths, magnitude-aware modeling remains future work.

For genetic perturbations, the initial condition reflects the intervention modality: CRISPRi uses $\mathrm{sign}_{0}=-1$ and $z_0=0$, whereas CRISPRa uses $\mathrm{sign}_{0}=+1$ and $z_0=1$. The remaining path extraction, sign propagation, and filtering procedures are identical across these genetic modalities.

\begin{algorithm}[ht]
\caption{Sign-Reachability Filter for Differential Expression}
\label{alg:reachability}
\SetAlgoLined
\KwIn{Extracted path $\pi=(a_1,\dots,a_L)$, Initial perturbation sign $\mathrm{sign}_{0}(p_i)$, Observed label $\ell_i \in \{\texttt{up}, \texttt{down}\}$, Ambiguity budget $B_0$}
\KwOut{Validity of the path (True/False)}
$\sS_0 \gets \{\mathrm{sign}_{0}(p_i)\}$\;
$b_{count} \gets 0$\;
\For{$t \gets 1$ \KwTo $L$}{
    \uIf{$\mathrm{sign}(a_t) == +1$}{
        $\sS_t \gets \sS_{t-1}$\;
    }
    \uElseIf{$\mathrm{sign}(a_t) == -1$}{
        $\sS_t \gets \{-s : s \in \sS_{t-1}\}$\;
    }
    \Else{
        $\sS_t \gets \sS_{t-1} \cup \{-s : s \in \sS_{t-1}\}$\;
        $b_{count} \gets b_{count} + 1$\;
    }
}
\If{$b_{count} > B_0$}{
    \Return False \tcp*{Path exceeds allowable ambiguity}
}
$target\_sign \gets +1$ \textbf{if} $\ell_i == \texttt{up}$ \textbf{else} $-1$\;
\Return $target\_sign \in \sS_L$\;
\end{algorithm}

\begin{algorithm}[ht]
\caption{Sign-Consistency Filter for Unchanged Targets}
\label{alg:consistency}
\SetAlgoLined
\KwIn{Extracted path $\pi=(a_1,\dots,a_L)$, Perturbation initial state $z_0 \in \{0, 1\}$, Target basal expression $x_{y_i,c_i}$, Thresholds $\tau_{\mathrm{low}}, \tau_{\mathrm{high}}$}
\KwOut{Consistency of the path (True/False)}
\uIf{$x_{y_i,c_i} \leq \tau_{\mathrm{low}}$}{
    $b \gets 0$ \tcp*{Basally silent}
}
\uElseIf{$x_{y_i,c_i} \geq \tau_{\mathrm{high}}$}{
    $b \gets 1$ \tcp*{Basally saturated}
}
\Else{
    $b \gets \varnothing$ \tcp*{Active state without boundaries}
}
$\sZ_0 \gets \{z_0\}$\;
\tcp{Propagate binary activity states}
\For{$t \gets 1$ \KwTo $L$}{
    \uIf{$\mathrm{sign}(a_t) == +1$}{
        $\sZ_t \gets \sZ_{t-1}$\;
    }
    \uElseIf{$\mathrm{sign}(a_t) == -1$}{
        $\sZ_t \gets \{1 - z : z \in \sZ_{t-1}\}$\;
    }
    \Else{
         $\sZ_t \gets \{0, 1\}$\;
    }
}
\tcp{Evaluate physiological constraints}
\If{$(b == 1 \land 1 \in \sZ_L) \lor (b == 0 \land 0 \in \sZ_L)$}{
    \Return True \tcp*{Ceiling or floor effect observed}
}
\Return False\;
\end{algorithm}

\subsubsection{Path-to-Text Reasoning Synthesis}

To systematically translate structured causal graphs into natural language, we implement a multi-stage prompt construction and generation pipeline. This process ensures that the resulting text maintains strict fidelity to the underlying biological mechanisms and properly grounds the language model in the defined cellular context.

For a given cell state $c$, the generation module requires the perturbation $p$, the effect gene $g_e$, the observed differential expression label $y \in \{\text{unchanged}, \text{up}, \text{down}\}$, and the fixed, filtered causal paths. To present this graph-structured data to the language model, we serialize the pathways using a flat breadth-first search traversal initialized at the perturbation root $p$. This topological ordering ensures that upstream regulatory events are structurally prioritized in the prompt before the downstream interactions that converge on $g_e$.

To provide the language model with an accurate representation of intermediate regulator availability, we discretize the continuous basal expression values of the genes along the selected paths. Leveraging the adaptive hysteresis thresholds $\tau_{\mathrm{low}}$ and $\tau_{\mathrm{high}}$ defined previously, we map the basal expression of each relevant gene into explicit qualitative states of low, medium, or high availability. This contextual translation allows the model to correctly deduce whether a specific signaling route is mechanistically viable, attenuated, or entirely insufficient to induce a change in $g_e$ within the initial cellular environment.

The comprehensive prompt integrates five core components: the cell type, the perturbation $p$, the effect gene $g_e$, the expanded text of the observed label $y$, and the serialized path evidence. We apply specific encoding rules for perturbations to maintain mechanistic integrity. The prompt instructs the model to synthesize a single predictive paragraph that derives $y$ purely as a logical consequence of the provided pathways, actively resolving ambiguous regulatory relations into defined polarities without relying on external or unsupported mechanisms.

Text generation is executed using Qwen3-4B configured with bfloat16 precision. For post-generation quality control, we identify recurring expressions of 3--10 words appearing in at least 10 reasoning texts and review their prevalence. From this analysis, we construct a fixed-phrase list that explicitly states the supplied pathway cannot explain or support the observed response, such as ``cannot be explained'', ``is not supported'', and ``not mechanistically supported''. We remove any reasoning text containing one or more of these phrases. This procedure is a conservative filter for explicit contradiction or non-support statements. Note that this filter does not independently establish biological validity or detect all hallucinations. The surviving samples are assigned deterministic identifiers based on the unique combination of cell type, perturbation, and effect gene, which are subsequently compiled into the final \PertReasonQA dataset.

\subsubsection{Hyperparameters}

In our implementation, the structural base weight $w^{\mathrm{base}}_{uv}$ was set to a uniform value of $1.0$. The gating steepness parameter $\beta$ was set to $10$. To maintain appropriate baseline traversability, the source floor $\delta_{\mathrm{src}}$ was configured to $0.05$, and the target floor $\delta_{\mathrm{tgt}}$ was configured to $0.5$. The small weight constant $\epsilon_{w} = 10^{-9}$ was added to the final weight computation to prevent exact zeros.

For the PageRank diffusion process, we utilized a damping parameter $\alpha = 0.85$, capped at a maximum of $50$ iterations with a tolerance of $10^{-4}$. In the path extraction phase, the edge cost formulation included a small constant $\epsilon_{c} = 10^{-4}$. The maximum hop-depth constraint $D_{\max}$ was restricted to $4$ (which is reduced to $3$ internally for chemical perturbations after prepending the drug-target edge). To construct the final Bag-of-Paths, the pipeline iteratively extracted up to $K=4$ diverse paths, where $K$ is the path budget, penalizing visited edges by a factor of $\lambda = 2.0$. The average pathway length in Table~\ref{tab:data_stats} summarizes the bag of paths for each reasoning sample, which can contain multiple paths.

For all path-to-text reasoning generations, we use Qwen's recommended configurations, specifically, temperature $0.7$, top-$p$ sampling $0.8$, and top-$k$ sampling $20$.

\section{Model and Training Details}\label{app:model}
This section describes the supervised fine-tuning curriculum and GRPO training procedure used to develop \PertReasonLM.

\subsection{SFT Training}\label{app:sft_impl}

We cast supervised fine-tuning as a standard causal language modeling objective. Let $x$ denote the input instruction and $y$ the target output. We minimize the negative log-likelihood over response tokens using a completion-only loss strategy, where only the assistant tokens contribute to the gradient:
\[
\mathcal{L}_{\text{SFT}} = -\sum_{t=1}^{|y|} \log P_{\theta}\!\left(y_t \mid x, y_{<t}\right),
\]
where $\theta$ represents the trainable Low-Rank Adaptation (LoRA) parameters.

\paragraph{Outcome-Only Supervision}

The primary objective of this initial stage is to enable \PertReasonLM to acquire domain knowledge regarding the broad associations among a perturbation $p$, a cellular context $c$, an effect gene $g_e$, and the expression outcome $y$. Because high-quality knowledge graph triplets for reasoning are scarce for many perturbations and effect gene pairs, outcome-only samples are significantly more abundant within \PertReasonQA. Consequently, this phase acts as a crucial knowledge acquisition step, allowing the model to internalize biological entity representations and perturbation response priors. By predicting the final outcome without requiring a full mechanistic explanation, this stage provides a stable initialization that primes the model for the subsequent, more complex reasoning supervision. In practice, the input sequence is constructed using the cell type, the perturbation $p$, the effect gene $g_e$, and the basal expression contexts. The corresponding output consists exclusively of the final differential expression label $y \in \{\text{unchanged, up, down}\}$, without any intermediate reasoning paths or retrieved pathway evidence.

\paragraph{Mechanistic Reasoning Supervision}

Following the initial knowledge acquisition phase, we train \PertReasonLM to generate Chain-of-Thought (CoT) reasoning in natural language~\citep{wei2022chain, chung2024scaling, magister2023teaching} that elucidates the causal pathways from the perturbation $p$ to the effect gene $g_e$. This supervision forces the model to align the ``broad associations" learned in the previous stage with structured, step-by-step ``causal mechanisms". By doing so, we mitigate shortcut predictions and ensure that the final concluded outcome $y$ is explicitly justified by established biological pathway evidence. Crucially, the model learns to jointly interpret retrieved pathway knowledge and the basal cell state $c$ to select the correct context-specific regulatory mechanism. We also employ data mixing that combines complex reasoning samples with the outcome-only samples~\citep{guo2025deepseek, mitra2023orca}. The target output is then formatted as a tag-structured response, including a \texttt{<thinking>} block of the mechanistic reasoning, a final \texttt{<answer>} block explicitly stating the predicted expression change $y$, and a \texttt{<triplet>} block detailing the used causal edges.

\paragraph{Relation Direction Self-Supervision}

Despite the integration of multiple KG sources, the precise regulatory direction of certain gene-gene interactions often remains unknown or ambiguous. To address this, the final stage of our SFT pipeline introduces a biologically inspired self-supervised learning objective. We formulate a ``relation masking and prediction'' task designed to teach the model edge-level mechanistic primitives, enabling it to accurately distinguish between activation and inhibition \citep{teru2020inductive, zhang2024making}. Here, the model is explicitly tasked with inferring the causal sign of ambiguous edges based on the cell and KG contexts. To ensure the model learns genuine context-based disambiguation rather than simply copying localized text, we mask the directionality of a subset of well-defined, clear relations during training. This strategy forces \PertReasonLM to actively use the provided cellular and pathway contexts to deduce the correct causal direction, thereby significantly reinforcing its foundational mechanistic understanding.

\paragraph{Multi-Stage SFT Curriculum}

The comprehensive SFT process is structured as a progressive curriculum that systematically advances from simpler task adaptations to mechanistically richer forms of supervision. Specifically, the training sequence unfolds across three phases: outcome-only adaptation, outcome-plus-reasoning alignment, and relation-direction specialization. Multi-objective consolidation is achieved through data mixing within this curriculum, integrating the three forms of supervision into a cohesive perturbation reasoning capability within \PertReasonLM. By carefully ordering these learning objectives, our curriculum effectively mitigates the inherent data trade-off in \PertReasonQA, bridging the gap between a large but shallow pool of outcome supervision and a small but high-fidelity set of mechanistic reasoning signals.

Due to the diverse nature of \PertReasonQA, the outcome-only, reasoning, and relation-direction samples exhibit significant disparities in both quantity and sequence length. Simple uniform batching across these distributions leads to objective and memory imbalances. To ensure stable optimization, we employ dynamic batch sampling. We construct homogeneous micro-batches containing samples from exclusively one supervision type and then mix these types at the gradient accumulation block level. This strategy maintains the required effective batch size for each specific objective type. Additionally, we apply oversampling based on the perturbation modality and outcome label to mitigate data skewness.

Training is implemented using the \texttt{TRL} library \citep{vonwerra2020trl} with DeepSpeed ZeRO-2 \citep{deepspeed} for memory efficiency on two NVIDIA A100 GPUs (40 GB each) with 128 GB system memory. All SFT stages share a common set of hyperparameters, which, along with stage-specific effective batch sizes, epochs, and masking ratios, are detailed in Table~\ref{tab:sft_hyperparams}.

\begin{table}[t]
\centering
\caption{Hyperparameters for Supervised Fine-Tuning}
\label{tab:sft_hyperparams}
\small
\begin{tabular}{lc}
\toprule
\textbf{Hyperparameter} & \textbf{Value} \\
\midrule
Precision & BFloat16 \\
Learning Rate & $2 \times 10^{-4}$ \\
LR Scheduler & Cosine \\
Warmup Ratio & 0.03 \\
Max Sequence Length & 4096 \\
Training Technique & Gradient Checkpointing, Completion-only Loss \\
LoRA Rank ($r$) & 32 \\
LoRA Alpha ($\alpha$) & 64 \\
LoRA Dropout & 0.05 \\
\midrule
\textbf{Outcome-Only Supervision} & \\
\quad Epoch & 1 \\
\quad Effective Batch Size & 512 \\
\textbf{Mechanistic Reasoning Supervision} &  \\
\quad Epoch & 1 \\
\quad Effective Batch Size & 128 \\
\textbf{Relation Direction Supervision} &  \\
\quad Epoch & 2 \\ 
\quad Effective Batch Size & 128 \\
\quad Clear Relation Masking Ratio & 20\% \\
\bottomrule
\end{tabular}
\end{table}

\begin{table}[t]
\centering
\small
\caption{GRPO implementation settings.}
\begin{tabularx}{\linewidth}{lX}
\toprule
\textbf{Component} & \textbf{Setting} \\
\midrule
Initialization & Pre-merged SFT checkpoint \\
Epoch & 3 \\
Rollouts per prompt & 8 completions \\
Sampling & Temperature $0.85$, top-$p=0.9$, max completion length $768$ \\
Objective & Clipped GRPO with KL coefficient $\beta=0.02$ against epoch-start reference \\
Reward & $0.8 \times r_{\mathrm{ans}} + 0.2 \times r_{\mathrm{trp}}$ \\
Optimization & AdamW, lr $5 \times 10^{-6}$, bf16, DeepSpeed ZeRO-2, gradient checkpointing \\
LoRA & Rank $16$, alpha $32$, dropout $0.05$; targets: \texttt{q/k/v/o\_proj}, \texttt{gate/up/down\_proj} \\
Stabilizers & GT injection into last rollout; zero-std rescue with baseline $\pm 1.0$ \\
\bottomrule
\end{tabularx}
\label{tab:rl_impl_details}
\end{table}

\subsection{GRPO Training}
\label{app:rl_impl}

\paragraph{Distributed Training Loop}
Each RL epoch runs as a five-phase workflow: (1)~\textbf{Prepare}: load JSONL data from cell-conditioned path context and no path context pools, apply two-stage oversampling, and append \texttt{/no\_think} to disable the backbone's native thinking mode. (2)~\textbf{Generate}: shard prompts across nodes and decode $K{=}8$ completions per prompt with vLLM; (3)~\textbf{Rewards}: optionally inject the ground-truth response as the final rollout (GT injection), compute scalar rewards, and normalize into group advantages; (4)~\textbf{Train}: attach a fresh LoRA adapter to the epoch-start merged checkpoint, compute reference log-probabilities with adapters disabled, and optimize with DeepSpeed ZeRO-2; (5)~\textbf{Merge}: fold the learned adapter back into the checkpoint for the next epoch.

Our GRPO runs on an 8-node cluster with 16 NVIDIA A100-40GB GPUs in total, and each epoch completes in under 10 hours. This hardware configuration supports the distributed rollout, reward, and DeepSpeed training pipeline used for the official model.

\paragraph{Advantage Normalization and Stabilization}
Group advantages are computed as
\[
A_i =
\begin{cases}
\dfrac{r_i - \mu_x}{\sigma_x}, & \sigma_x > 0, \\[4pt]
+1.0, & \sigma_x = 0 \text{ and } \mu_x > 0.5, \\[2pt]
-1.0, & \sigma_x = 0 \text{ and } \mu_x \leq 0.5.
\end{cases}
\]
The zero-variance rescue prevents zero-signal updates on groups where all completions receive identical rewards, which is common on structured biological outputs. GT injection into the last rollout further reduces the frequency of such groups.

\paragraph{Reward Design}

The reward balances outcome prediction with recovery of reference regulatory mechanisms: $r^{\mathrm{sel}} = 0.8\,r_{\mathrm{ans}} + 0.2\,r_{\mathrm{trp}}$. Here, $r_{\mathrm{ans}}$ is a binary indicator of final-answer correctness. The triplet reward $r_{\mathrm{trp}}$ measures partial recovery of reference interactions, assigning each interaction a score of 1 for a correct entity pair and regulatory sign, 0.4 for a correct pair with an unspecified sign, 0.25 for a correct pair with an incorrect sign, and 0 for a missing pair, then averaging over reference interactions. This graded supervision encourages recovery of regulatory structure while favoring correct directionality, providing a learning signal even when the final outcome prediction is incorrect.

\paragraph{Hyperparameters}
Full GRPO settings are listed in Table~\ref{tab:rl_impl_details}.

\section{Baseline Details} \label{app:baselines}
This section introduces the gene-space and text-space baselines and describes how they are evaluated on \PertReasonQA.

\subsection{Gene-Space Baselines with Numerical Outputs}\label{app:baselines_gene}

\begin{itemize}
    \item \textbf{GEARS}~\cite{roohani2024predicting}: A KG-based model for perturbation prediction that uses gene co-expression and Gene Ontology-derived similarity graphs to predict outcomes for unseen single and combinatorial genetic perturbations.

    \item \textbf{scGPT}~\cite{cui2024scgpt}: A generative transformer-based foundation model pre-trained on over 33 million single-cell transcriptomes to learn generalizable gene and cell representations. We fine-tune scGPT on our dataset to predict perturbation-induced gene-expression changes using its generative modeling framework.
    
    \item \textbf{STATE (SE+ST)}~\cite{adduri2026predicting}: A perturbation-focused foundation model that combines a State Embedding (SE) model, pre-trained on large-scale observational single-cell data, with a State Transition (ST) model trained on perturbation-response data to predict changes in gene expression across cell populations. We use embeddings from the pre-trained SE model and train the ST model from scratch on our dataset.

\end{itemize}

These baselines are trained on CRISPRi perturbations and evaluated only on the four matched CRISPRi genetic test folds. This is the fairest protocol because GEARS and scGPT consume gene-level inputs and predict continuous perturbed expression profiles, rather than operating over a representation that can natively encode both genes and small molecules.

For STATE, while genetic perturbations are represented via ESM2 protein embeddings, allowing the model to generalize to unseen genes via sequence similarity, its chemical perturbation module typically employs embedding lookup by one-hot encodings. This dependency on fixed vocabularies makes chemical OOD extrapolation technically impossible. Consequently, to ensure a fair comparison focused on generalization, we evaluate these numerical baselines exclusively on the matched CRISPRi genetic subset of \PertReasonQA.

In addition, we exclude the CRISPRa single-to-combo fold for GEARS, scGPT, and STATE because its intervention modality differs from their CRISPRi training.

\paragraph{Continuous-to-Label Conversion}

These baselines output continuous gene-expression values rather than 3-way labels. To compare them against our benchmark, we convert each predicted perturbation profile into differential-expression calls by pairing the predicted perturbed values with empirical control cells from the same data source and cell type, then performing a Wilcoxon rank-sum test for each $(c,p)$ contrast. We apply Benjamini-Hochberg correction across the genes available in that contrast and assign \texttt{up} or \texttt{down} only when the adjusted $p$-value passes the evaluation threshold and the estimated log-fold change has the corresponding sign; all remaining cases are mapped to \texttt{unchanged}. If a cell-type-specific control pool is unavailable, we fall back to the corresponding source-level controls. This protocol evaluates the induced 3-way endpoint rather than raw regression error, which is the quantity aligned with our QA benchmark.

\subsection{Text-Space Baselines Based on LLMs}

\begin{itemize}
    \item \textbf{Qwen3-4B (base)}~\cite{yang2025qwen3technicalreport}: The backbone of \PertReasonLM-4B, evaluated without any fine-tuning to isolate the gains from our training pipeline.

    \item \textbf{BioMistral 7B}~\cite{labrak2024biomistralcollectionopensourcepretrained} and \textbf{NatureLM 8x7B}~\cite{xia2025naturelanguagemodeldeciphering}: Leading open-source biomedical LLMs continuously pre-trained or instruction-tuned on large-scale biomedical corpora (PubMed, medical guidelines). They represent the upper bound of performance for models relying solely on implicit pre-training knowledge without explicit KG-guided reasoning.

    \item \textbf{SUMMER}~\cite{wu2025perturbqa}: A state-of-the-art Retrieval-Augmented Generation baseline. It retrieves relevant literature or KG triplets to answer perturbation queries, serving as a strong RAG reference point. We use Qwen3-4B as a backbone model for SUMMER.
\end{itemize}

\section{Limitations of Gene-Centric Baselines for Chemical Perturbations}\label{app:chemical_baseline_limitations}

GEARS and scGPT are designed to predict transcriptional responses to genetic perturbations and do not directly accept small-molecule drugs as perturbation inputs. One possible workaround is to map each drug to its known protein targets using INDRA~\cite{indra} and represent these targets as a combinatorial genetic perturbation. However, the GEARS and scGPT implementations used in our experiments support at most two perturbed genes per input. Consequently, a drug with more than two known targets cannot be fully represented through this approach: only two targets can be retained, while the remaining targets must be discarded.

This restriction differs from PertReasonLM, which represents chemical perturbations through textual descriptions and knowledge-grounded drug-target interactions. Rather than restricting each drug to two target genes, PertReasonLM can incorporate a variable number of known targets into its pathway context, allowing it to reason over multiple potential mechanisms connecting a drug to a downstream expression response.

We explored the chemical-to-genetic proxy by mapping each drug to two selected protein targets from INDRA~\cite{indra} and providing them as a combinatorial genetic perturbation to the gene-space models. However, this approximation introduces two limitations:

\begin{enumerate}
    \item \textbf{Incomplete drug-target representation.} Drugs can interact with multiple proteins, but the two-gene input limit in the evaluated GEARS and scGPT implementations forces the omission of additional known targets. This can remove biologically relevant mechanisms contributing to the observed transcriptional response.
    
    \item \textbf{Mismatch between chemical and genetic perturbations.} Even when all relevant targets could be included, representing drug-target interactions as genetic interventions does not capture differences in binding affinity, partial inhibition, or other pharmacological effects. Thus, a combinatorial genetic perturbation is not necessarily equivalent to treatment with the corresponding drug.
\end{enumerate}

STATE presents a different limitation. Although it supports chemical perturbations, the chemical embedding lookup used in our configuration relies on a fixed vocabulary and does not support unseen compounds. Therefore, its chemical representation is unsuitable for the evaluation of OOD chemical perturbations considered here.

The chemical-to-genetic proxy produced high-variance predictions, but these results are confounded by the incomplete target representation and the mismatch between chemical and genetic interventions. We therefore exclude proxy-based chemical results from the main comparison. Instead, GEARS, scGPT, and STATE are trained and evaluated on the four matched CRISPRi genetic test folds.

\section{Results of GPT Families on \PertReasonQA-mini}\label{app:mini_benchmark}
To make API-based evaluation more tractable, we construct \PertReasonQA-mini by randomly subsampling the full benchmark so that each of the ten test slices contains at most 200 examples.
The reported total mini-benchmark size is 1,991 across the ten test folds.
This preserves the benchmark split structure while reducing the cost of closed-source evaluation; in our setup, a full GPT-5.4 run on \PertReasonQA-mini costs roughly \$15, while GPT-5.6 Sol costs around \$23.
The GPT model API setting used in this work did not provide internet search, browsing, or external retrieval tools, so these models answered from the prompt and their parametric knowledge alone.

\begin{table}[t]
\centering
\small
\caption{Results on \PertReasonQA-full and \PertReasonQA-mini, macro-averaged across ten test folds.}
\shrinktable{\linewidth}{%
\begin{tabular}{llccc}
\toprule
Dataset & Model & Balanced Accuracy & Edge Recall & GO Similarity \\
\midrule
\PertReasonQA-full & SUMMER              & 0.397 & 0.818 & 0.930 \\
\PertReasonQA-full & Qwen3-4B            & 0.364 & 0.840 & 0.886 \\
\PertReasonQA-full & \PertReasonLM-SFT       & 0.681 & 0.979 & 0.938 \\
\PertReasonQA-full & \PertReasonLM-GRPO      & \textbf{0.704} & \textbf{0.979} & \textbf{0.938} \\
\midrule
\PertReasonQA-mini & SUMMER         & 0.391  & 0.814  & 0.928  \\
\PertReasonQA-mini & Qwen3-4B       & 0.371  & 0.837  & 0.884  \\
\PertReasonQA-mini & \PertReasonLM-SFT  & 0.679  & 0.979  & 0.937  \\
\PertReasonQA-mini & \PertReasonLM-GRPO & \textbf{0.708}  & \textbf{0.980}  & \textbf{0.938}  \\
\bottomrule
\end{tabular}%
}
\label{tab:appendix_mini_alignment}
\end{table}

\begin{table}[t]
\caption{Balanced accuracy by split and pathway context on \PertReasonQA-mini. Average gives the equally weighted mean across the ten splits.}
\label{tab:appendix_mini_benchmark}
\centering
\small
\setlength{\tabcolsep}{3pt}
\shrinktable{\linewidth}{%
\begin{tabular}{@{}l*{11}{c}@{}}
\toprule
 & & \multicolumn{4}{c}{ID Cell} & \multicolumn{4}{c}{OOD Cell} & & \\
\cmidrule(lr){3-10}
\multirow{-2}{*}{Model} & \multirow{-2}{*}{Average} & ID Chem. & ID Gen. & OOD Chem. & OOD Gen. & ID Chem. & ID Gen. & OOD Chem. & OOD Gen. & \multirow{-2}{*}{\begin{tabular}[c]{@{}c@{}}Gen-to\\ Chem\end{tabular}} & \multirow{-2}{*}{\begin{tabular}[c]{@{}c@{}}Single-to\\ Combo\end{tabular}} \\
\midrule
\multicolumn{12}{l}{\textbf{Cell-conditioned paths}} \\
\midrule
Qwen3-4B base & 0.371 & 0.416 & 0.337 & 0.380 & 0.346 & 0.417 & 0.371 & 0.320 & 0.358 & 0.378 & 0.388 \\
GPT-5.4 & 0.386 & 0.398 & 0.363 & 0.377 & 0.368 & 0.364 & 0.415 & 0.391 & 0.427 & 0.344 & 0.411 \\
GPT-5.6 Sol & 0.437 & 0.461 & 0.407 & 0.440 & 0.385 & 0.429 & 0.493 & 0.441 & 0.515 & 0.395 & 0.402 \\
\PertReasonLM-SFT & 0.679 & 0.719 & 0.726 & \textbf{0.694} & 0.689 & 0.642 & 0.742 & 0.673 & 0.771 & \textbf{0.631} & 0.508 \\
\PertReasonLM-GRPO & \textbf{0.708} & \textbf{0.742} & \textbf{0.753} & 0.685 & \textbf{0.771} & \textbf{0.729} & \textbf{0.778} & \textbf{0.694} & \textbf{0.803} & 0.573 & \textbf{0.550} \\
\midrule
\multicolumn{12}{l}{\textbf{No paths}} \\
\midrule
Qwen3-4B base & 0.338 & 0.411 & 0.348 & 0.353 & 0.284 & 0.435 & 0.342 & 0.303 & 0.327 & 0.283 & 0.294 \\
GPT-5.4 & 0.405 & 0.432 & 0.458 & 0.435 & 0.413 & 0.438 & 0.438 & 0.392 & 0.353 & 0.285 & 0.405 \\
GPT-5.6 Sol & 0.440 & 0.457 & 0.483 & 0.463 & 0.444 & 0.483 & 0.460 & 0.436 & 0.405 & 0.375 & 0.390 \\
\PertReasonLM-SFT & 0.672 & 0.720 & 0.696 & 0.674 & 0.699 & \textbf{0.721} & 0.732 & 0.662 & 0.727 & \textbf{0.556} & 0.537 \\
\PertReasonLM-GRPO & \textbf{0.701} & \textbf{0.783} & \textbf{0.763} & \textbf{0.676} & \textbf{0.749} & 0.715 & \textbf{0.761} & \textbf{0.669} & \textbf{0.764} & 0.542 & \textbf{0.586} \\
\bottomrule
\end{tabular}%
}
\end{table}

\begin{table}[t]
\caption{Aggregate performance by pathway context on \PertReasonQA-mini. Edge recall, path connectivity, and GO similarity summarize reasoning quality.}
\label{tab:appendix_mini_reasoning}
\centering
\small
\setlength{\tabcolsep}{4pt}
\begin{tabular}{lcccc}
\toprule
Model & Balanced Accuracy & Edge Recall & Path connectivity & GO similarity \\
\midrule
\multicolumn{5}{l}{\textbf{Cell-conditioned paths}} \\
\midrule
Qwen3-4B base & 0.371 & 0.837 & 0.987 & 0.884 \\
GPT-5.4 & 0.386 & 0.750 & 0.899 & 0.849 \\
GPT-5.6 Sol & 0.437 & 0.823 & 0.989 & 0.868 \\
\PertReasonLM-SFT & 0.679 & 0.979 & 0.980 & 0.937 \\
\PertReasonLM-GRPO & \textbf{0.708} & \textbf{0.980} & \textbf{0.995} & \textbf{0.938} \\
\midrule
\multicolumn{5}{l}{\textbf{No paths}} \\
\midrule
Qwen3-4B base & 0.338 & 0.137 & 0.864 & 0.615 \\
GPT-5.4 & 0.405 & 0.198 & 0.605 & 0.611 \\
GPT-5.6 Sol & 0.440 & 0.225 & 0.689 & 0.661 \\
\PertReasonLM-SFT & 0.672 & 0.341 & \textbf{0.991} & 0.750 \\
\PertReasonLM-GRPO & \textbf{0.701} & \textbf{0.396} & 0.987 & \textbf{0.778} \\
\bottomrule
\end{tabular}%
\end{table}

Table~\ref{tab:appendix_mini_alignment} supports using \PertReasonQA-mini for GPT comparisons: balanced accuracy differs from the full benchmark by at most 0.7 percentage points across shared models, with their ranking preserved.

Table~\ref{tab:appendix_mini_benchmark} shows that both of our models substantially outperform GPT-5.4 and GPT-5.6 Sol. Under the cell-conditioned path context, \PertReasonLM-SFT and \PertReasonLM-GRPO achieve balanced accuracies of 0.679 and 0.708, versus 0.386 for GPT-5.4 and 0.437 for GPT-5.6 Sol. Under the no path context, SFT and GRPO retain advantages of 23.2 and 26.1 percentage points over the stronger GPT-5.6 Sol baseline. Both of our models outperform both GPT models on every split under both settings.

Table~\ref{tab:appendix_mini_reasoning} shows corresponding gains in reasoning quality. With the cell-conditioned path context, SFT and GRPO achieve edge recall of 0.979/0.980 and GO similarity of 0.937/0.938, compared with 0.823 and 0.868 for GPT-5.6 Sol. With the no path context, GRPO reaches 0.396 edge recall, 0.987 path connectivity, and 0.778 GO similarity, versus 0.225, 0.689, and 0.661 for GPT-5.6 Sol; SFT also exceeds both GPT models on all three metrics. Thus, the advantages over GPT extend beyond outcome prediction to recovery of the reference mechanisms.

\section{BioMistral and NatureLM: Prediction-Only Fallback}\label{app:biomed_llm_failures}
\begingroup
\definecolor{biomedquestion}{RGB}{36,67,88}
\definecolor{biomedoutput}{RGB}{111,70,43}
\lstdefinestyle{biomedfailure}{
  style=wrappedtextbox,
  basicstyle=\ttfamily\footnotesize,
  breakatwhitespace=true,
  frame=single,
  framerule=0.4pt,
  rulecolor=\color{biomedoutput!40},
  backgroundcolor=\color{biomedoutput!3},
  framesep=7pt,
  xleftmargin=8pt,
  xrightmargin=8pt,
  aboveskip=8pt,
  belowskip=4pt
}
\newenvironment{biomedfailureexample}
  {\par\medskip\noindent\begin{minipage}{\linewidth}\small}
  {\end{minipage}\par\medskip}

We first evaluated BioMistral and NatureLM under the same structured reasoning protocol used for the main benchmark, requiring a valid \texttt{<thinking>}, \texttt{<answer>}, and \texttt{<triplet>} output. In practice, both models failed to produce reliable structured generations often enough that the full chain-of-thought evaluation was not informative. BioMistral was the more extreme case, failing to emit a parseable structured answer on {97\%} of samples. NatureLM produced more text, but much of it drifted away from the task format or ignored the supplied context graph.

To obtain a fairer lower-bound comparison, we re-ran both models in a prediction-only setting and evaluated only the final label. Even under this relaxed protocol, both models remained close to chance because their predictions were highly skewed across the three labels. BioMistral never predicted the up class in either context, yielding balanced accuracy of roughly {31.3\%} overall despite moderate recall on down. NatureLM showed a milder but still severe skew in the no path context, with macro-averaged recall of {14.0\%} on up, {66.6\%} on down, and {16.1\%} on unchanged, corresponding to {32.2\%} balanced accuracy. In the cell-conditioned path context, NatureLM effectively collapsed toward predicting down, with macro-averaged recall of {98.6\%} on down, {1.0\%} on up, and {0.5\%} on unchanged, giving {33.4\%} balanced accuracy.

The main takeaway is that domain-specific pretraining alone does not solve this task. Without perturbation-specific supervision and format control, these biomedical LLMs fail twice: first at structured mechanistic generation, and second at balanced three-way label prediction even after the reasoning requirement is removed.

We also qualitatively examine why BioMistral and NatureLM could not be meaningfully compared under the full structured reasoning protocol. We group the failed generations into three recurring behavioral modes.

\subsection{Failure Mode 1: Degenerative Repetition}
NatureLM falls into infinite loops of biological terms or path logic, failing to synthesize a final conclusion.

\begin{biomedfailureexample}
\textbf{\color{biomedquestion}Question.}\par\nobreak\smallskip\noindent
Is the expression of CENPU up, down, or unchanged after genetic perturbation
CRISPRi knockdown of CXXC1 in rpe1? ...
\par\smallskip
\noindent\textbf{NatureLM Output.}\par\nobreak
\begin{lstlisting}[style=biomedfailure]
- CXXC1 activates RB1 => CDK1 => CENPU
- RB1 activates JUN => NR3C1 => CENPU
- CXXC1 activates RB1 => CDK1 => CENPU
- RB1 activates JUN => NR3C1 => CENPU
... [Repeats Until Token Limit]
\end{lstlisting}
\end{biomedfailureexample}

\subsection{Failure Mode 2: Context Ignorance and Encyclopedic Drift}
Rather than reasoning over the provided context-specific graph, models substitute generic training data.

\par\medskip\noindent\textbf{Sub-mode 2a: Metadata retrieval.}\par\nobreak
\begin{biomedfailureexample}
\textbf{\color{biomedquestion}Question.}\par\nobreak\smallskip\noindent
Is the expression of AMOT up, down, or unchanged after chemical perturbation
Mocetinostat (MGCD0103) in a549? ...
\par\smallskip
\noindent\textbf{NatureLM Output.}\par\nobreak
\begin{lstlisting}[style=biomedfailure]
The data is available on GEO under accession number: GSE139947?
\end{lstlisting}
\end{biomedfailureexample}

\par\medskip\noindent\textbf{Sub-mode 2b: Literal echoing.}\par\nobreak
\begin{biomedfailureexample}
\textbf{\color{biomedquestion}Question.}\par\nobreak\smallskip\noindent
Is the expression of CCL5 up, down, or unchanged after chemical perturbation
Vorinostat in T cells? ...
\par\smallskip
\noindent\textbf{BioMistral Output.}\par\nobreak
\begin{lstlisting}[style=biomedfailure]
"up", "down", or "unchanged".
\end{lstlisting}
\end{biomedfailureexample}

\subsection{Failure Mode 3: Instruction and Template Leakage}
BioMistral outputs the underlying conditional logic of the evaluation template rather than applying it.

\begin{biomedfailureexample}
\textbf{\color{biomedquestion}Question.}\par\nobreak\smallskip\noindent
Is the expression of HELB up, down, or unchanged after chemical perturbation
Bosutinib in T cells? ...
\par\smallskip
\noindent\textbf{BioMistral Output.}\par\nobreak
\begin{lstlisting}[style=biomedfailure]
If the answer is unchanged, then explain how the retrieved knowledge indicates
that the effect of the perturbation is not propagated to the gene of interest.
If the answer is changed, then explain how the retrieved knowledge indicates
that the effect of the perturbation is propagated to the gene of interest.
\end{lstlisting}
\end{biomedfailureexample}

\endgroup

\section{Training-Budget Controls for Mechanistic SFT}
\label{app:budget_controls}

To assess the contribution of mechanistic supervision beyond additional training, we extend Outcome SFT with two outcome-only continuations. Both start from the same Outcome-SFT checkpoint. The sample-matched continuation meets or slightly exceeds the CoT SFT sample budget, while the token-matched continuation meets or slightly exceeds its token budget. We compare these controls with the original Outcome SFT and CoT SFT results from Table~\ref{tab:main_path}.

\begin{table}[t]
\centering
\small
\caption{Training-budget controls for \PertReasonLM-4B, evaluated by balanced accuracy across eight test conditions under the cell-conditioned path context. Both outcome-only continuations start from the same Outcome-SFT checkpoint. Original rows reproduce the corresponding results.}
\label{tab:budget_controls}
\setlength{\tabcolsep}{3pt}
\shrinktable{\linewidth}{%
\begin{tabular}{@{}l*{10}{c}@{}}
\toprule
 & \multicolumn{2}{c}{Training budget} & \multicolumn{4}{c}{ID Cell} & \multicolumn{4}{c}{OOD Cell} \\
\cmidrule(lr){2-3}\cmidrule(lr){4-7}\cmidrule(lr){8-11}
Model & Samples & Tokens & \makecell{ID\\Chem.} & \makecell{ID\\Gen.} & \makecell{OOD\\Chem.} & \makecell{OOD\\Gen.} & \makecell{ID\\Chem.} & \makecell{ID\\Gen.} & \makecell{OOD\\Chem.} & \makecell{OOD\\Gen.} \\
\midrule
\makecell[l]{Outcome SFT\\(original)} & 3.0M & 631M & 0.713 & 0.690 & 0.612 & 0.622 & 0.653 & 0.646 & 0.543 & 0.611 \\
\addlinespace
\makecell[l]{Outcome SFT\\(sample matched)} & 3.8M & 807M & 0.719 & 0.708 & 0.639 & 0.666 & 0.675 & 0.661 & 0.552 & 0.601 \\
\addlinespace
\makecell[l]{Outcome SFT\\(token matched)} & 4.7M & 982M & 0.747 & 0.687 & 0.631 & 0.562 & 0.673 & 0.655 & 0.541 & 0.593 \\
\addlinespace
\makecell[l]{CoT SFT\\(original)} & 3.7M & 979M & \textbf{0.749} & \textbf{0.722} & \textbf{0.678} & \textbf{0.708} & \textbf{0.705} & \textbf{0.697} & \textbf{0.669} & \textbf{0.743} \\
\bottomrule
\end{tabular}%
}
\end{table}

Table~\ref{tab:budget_controls} shows that CoT SFT outperforms both outcome-only continuations on all eight splits. The largest gains occur under joint cell-and-perturbation shifts. On the two joint-OOD conditions, CoT SFT reaches 0.669 and 0.743, compared with 0.552 and 0.601 for the sample-matched continuation. Additional outcome-only samples or tokens therefore do not reproduce the benefit of the full mechanistic SFT stage. The comparison evaluates the full mechanistic SFT stage and does not isolate the contributions of individual components.

\section{Answer--Reason Error Taxonomy}\label{app:error_taxonomy}

This appendix formalizes the four-level answer--reason taxonomy used in the main analysis. 
The taxonomy directly crosses final-answer correctness with mechanistic correctness. 
Answer correctness is determined by comparing the final prediction extracted from the last valid \texttt{<answer>} block against the ground-truth perturbation outcome. 
Mechanistic correctness is determined from the canonicalized signed regulatory triplets extracted from the \texttt{<triplet>} block. 
A reason is marked correct only when triplets exist, the predicted triplets form a directed connected path from the perturbation source to the queried effect gene, and edge recall is at least $0.5$.

\begin{table}[t]
\centering
\caption{Four-level answer--reason taxonomy used throughout the analysis. The categories cross final-answer correctness with mechanistic correctness.}
\label{tab:answer_reason_taxonomy}
\small
\begin{tabularx}{0.92\textwidth}{@{}lX@{}}
\toprule
\textbf{Label} & \textbf{Rule} \\
\midrule
\texttt{Ans $\checkmark$ Reason $\checkmark$} 
& answer correct; triplets exist; path connected; edge recall $\geq 0.5$ \\

\texttt{Ans $\checkmark$ Reason $\times$} 
& answer correct; no triplets, disconnected path, or edge recall $< 0.5$ \\

\texttt{Ans $\times$ Reason $\checkmark$} 
& answer wrong; triplets exist; path connected; edge recall $\geq 0.5$ \\

\texttt{Ans $\times$ Reason $\times$} 
& answer wrong; no triplets, disconnected path, or edge recall $< 0.5$ \\
\bottomrule
\end{tabularx}
\end{table}

\begin{table}[t]
\centering
\caption{Four-level answer--reason counts under the cell-conditioned path context. Raw counts and model-to-model deltas for Qwen3-4B base, \PertReasonLM-SFT, and \PertReasonLM-GRPO cover all ten test folds (29,477 examples per model). Deltas are net category-count differences.}
\label{tab:error_shift_noisy}
\small
\shrinktable{\textwidth}{%
\begin{tabular}{@{}lrrrrr@{}}
\toprule
& \multicolumn{3}{c}{\textbf{Model Raw Counts}} 
& \multicolumn{2}{c}{\textbf{Performance Deltas}} \\
\cmidrule(lr){2-4} \cmidrule(l){5-6}
\textbf{Label} 
& \textbf{Qwen Base} 
& \textbf{SFT} 
& \textbf{GRPO} 
& \textbf{Base$\rightarrow$SFT} 
& \textbf{SFT$\rightarrow$GRPO} \\
\midrule
\texttt{Ans $\checkmark$ Reason $\checkmark$} 
& 9{,}776 & 20{,}454 & 21{,}393 & +10{,}678 & +939 \\

\texttt{Ans $\checkmark$ Reason $\times$} 
& 837 & 836 & 339 & -1 & -497 \\

\texttt{Ans $\times$ Reason $\checkmark$} 
& 16{,}470 & 7{,}995 & 7{,}664 & -8{,}475 & -331 \\

\texttt{Ans $\times$ Reason $\times$} 
& 2{,}394 & 192 & 81 & -2{,}202 & -111 \\
\bottomrule
\end{tabular}%
}
\end{table}

Table~\ref{tab:error_shift_noisy} reports the full four-level taxonomy counts. From Qwen3-4B base to \PertReasonLM-SFT, fully correct predictions increase from 9{,}776 to 20{,}454, a gain of 10{,}678 cases. At the same time, \texttt{Ans $\times$ Reason $\checkmark$} decreases from 16{,}470 to 7{,}995, and \texttt{Ans $\times$ Reason $\times$} decreases from 2{,}394 to 192. This pattern shows improvements in both answer correctness and the recovery of connected, reference-overlapping mechanisms.
GRPO adds a smaller but targeted refinement. It increases \texttt{Ans $\checkmark$ Reason $\checkmark$} by another 939 cases, reduces \texttt{Ans $\checkmark$ Reason $\times$} from 836 to 339, and further lowers both wrong-answer categories. Thus, the RL stage sharpens answer--reason agreement rather than qualitatively changing the overall failure profile.

\section{Biological Validation of Reference Mechanisms}
\label{app:reference_audit}
This section presents the objective, review protocol, and findings of a literature-based biological audit of \PertReasonQA reference mechanisms.

\subsection{Audit Objective}
\PertReasonQA retains KG-derived reference paths whose signed propagation and basal-state compatibility agree with observed perturbation outcomes. We conducted a post hoc audit of frozen references to assess their biological support through literature review. The audit examined evidence for critical interactions, complete paths, and the queried cellular context.

\subsection{Audit Design and Criteria}
The audit comprised 160 randomly sampled references, with 40 in each of four settings: ID cell / ID perturbation, OOD cell / ID perturbation, ID cell / OOD perturbation, and OOD cell / OOD perturbation. The same sample included 80 references each from chemical and genetic perturbations. Each review packet contained the cell type, perturbation, effect gene, observed outcome, complete candidate path, proximal and terminal critical edges, and evidence leads. A biology expert with single-cell/pathology expertise reviewed supporting and contradictory literature and recorded PubMed identifiers, PubMed Central identifiers, and digital object identifiers where available.
The review covered path-level judgments for all 160 references, 320 targeted retrieval queries for proximal and terminal edges, and 739 edge-level checks. These checks covered all critical edges and every edge in a prioritized subset of 100 paths.

\textbf{Critical-edge support} records evidence for both proximal and terminal anchors or at least one anchor. \textbf{Every-required-edge support} records evidence for all indispensable path edges. \textbf{Complete-path status} classifies a path as supported, uncertain, or unsupported through a joint assessment of edge support, sign consistency, chain coherence, cell-context compatibility, outcome support, contradictory evidence, and confidence. \textbf{Contradictory evidence} and \textbf{exact-cell/context evidence} record contrary findings and evidence matching the queried cell or context, respectively. Together, these criteria distinguish support for individual interactions, the overall mechanism, and its cellular context.

\subsection{Results and Interpretation}
Table~\ref{tab:reference_audit} summarizes the audit. At least one critical edge had support in 91.2\% of references. Complete-path judgments classified 58.1\% as supported, 38.8\% as uncertain, and 3.1\% as unsupported. Every-required-edge support was recorded for 26.2\% of the full sample, and exact-cell/context evidence was available for 26.9\%. Contradictory evidence was identified in 2.5\%. These findings show frequent support for pathway anchors alongside greater uncertainty about complete mechanisms and their cellular context.

\begin{table}[tbp]
\centering
\small
\caption{Biological evidence audit of 160 frozen \PertReasonQA references. The same 160 references are summarized overall, by perturbation modality, and by evaluation setting. Each reference contributes to one modality column and one evaluation-setting column. Entries are counts and within-column percentages. Path-level judgments covered all 160 references; exhaustive edge checks covered a prioritized subset of 100 paths.}
\label{tab:reference_audit}
\setlength{\tabcolsep}{4pt}
\renewcommand{\arraystretch}{1.15}
\shrinktable{\linewidth}{%
\begin{tabular}{@{}l*{7}{c}@{}}
\toprule
\multirow{2}{*}{Audit outcome} & \multirow{2}{*}{Overall} & \multicolumn{2}{c}{By perturbation modality} & \multicolumn{4}{c}{By evaluation setting} \\
\cmidrule(lr){3-4}\cmidrule(lr){5-8}
 & & Chemical & Genetic & \makecell{ID cell /\\ID perturbation} & \makecell{OOD cell /\\ID perturbation} & \makecell{ID cell /\\OOD perturbation} & \makecell{OOD cell /\\OOD perturbation} \\
\midrule
\makecell[l]{Critical edges:\\both supported} & 83 (51.9\%) & 53 (66.2\%) & 30 (37.5\%) & 22 (55.0\%) & 17 (42.5\%) & 22 (55.0\%) & 22 (55.0\%) \\
\makecell[l]{Critical edges:\\at least one supported} & 146 (91.2\%) & 75 (93.8\%) & 71 (88.8\%) & 36 (90.0\%) & 37 (92.5\%) & 35 (87.5\%) & 38 (95.0\%) \\
\makecell[l]{Every required\\edge supported} & 42 (26.2\%) & 27 (33.8\%) & 15 (18.8\%) & 12 (30.0\%) & 12 (30.0\%) & 14 (35.0\%) & 4 (10.0\%) \\
Path supported & 93 (58.1\%) & 58 (72.5\%) & 35 (43.8\%) & 27 (67.5\%) & 17 (42.5\%) & 28 (70.0\%) & 21 (52.5\%) \\
Path uncertain & 62 (38.8\%) & 18 (22.5\%) & 44 (55.0\%) & 13 (32.5\%) & 22 (55.0\%) & 8 (20.0\%) & 19 (47.5\%) \\
Path unsupported & 5 (3.1\%) & 4 (5.0\%) & 1 (1.2\%) & 0 (0.0\%) & 1 (2.5\%) & 4 (10.0\%) & 0 (0.0\%) \\
\makecell[l]{Any contradictory\\evidence} & 4 (2.5\%) & 4 (5.0\%) & 0 (0.0\%) & 0 (0.0\%) & 1 (2.5\%) & 3 (7.5\%) & 0 (0.0\%) \\
\makecell[l]{Exact-cell/context\\evidence} & 43 (26.9\%) & 33 (41.2\%) & 10 (12.5\%) & 18 (45.0\%) & 11 (27.5\%) & 6 (15.0\%) & 8 (20.0\%) \\
\midrule
Number of references & 160 & 80 & 80 & 40 & 40 & 40 & 40 \\
\bottomrule
\end{tabular}%
}
\end{table}

Chemical references had higher path support and more exact-context evidence than genetic references, which were more often uncertain. Chemical references also accounted for four of the five unsupported paths and all four cases with contradictory evidence. Across splits, path support was lowest in the OOD cell / ID perturbation setting and highest in the ID cell / OOD perturbation setting. The OOD cell / OOD perturbation setting combined frequent critical-edge support with limited every-required-edge support (4/40), illustrating the value of assessing these dimensions separately. These comparisons describe the audited sample.

The audit supports interpreting \PertReasonQA references as evidence-grounded mechanistic proxies with explicit uncertainty. Reference agreement measures recovery of the selected regulatory route, with interpretation guided by its biological support and context specificity. Alternative biologically plausible routes may receive lower edge-overlap scores.

\section{Trace Diagnostics and Metric Stress Tests}
\label{app:metric_diagnostics}

This section examines generated trace properties and tests how automatic metrics respond to controlled changes in reference mechanisms.

\subsection{Trace Diagnostics}

Table~\ref{tab:trace_diagnostics} reports 8-split mean predicted edges, strict edge precision, recall, F1, functional similarity, and outcome balanced accuracy on the \PertReasonQA-mini evaluation. Strict edge matching compares the source, relation, and target of canonicalized signed triplets. Precision measures the fraction of predicted edges matching the reference, recall measures reference-edge coverage, and F1 is their harmonic mean. Mean predicted edges summarizes the number of parsed edges per trace.

\begin{table}[t]
\centering
\small
\caption{Trace diagnostics on \PertReasonQA-mini evaluation under the cell-conditioned path context, with all metrics macro-averaged across eight test folds (without genetic-to-chemical and single-to-combo). Precision, recall, and F1 use strict reference-edge matching; mean edges is the average number of parsed predicted triplets per example.}
\label{tab:trace_diagnostics}
\setlength{\tabcolsep}{3pt}
\renewcommand{\arraystretch}{1.1}
\shrinktable{\linewidth}{%
\begin{tabular}{@{}l*{6}{c}@{}}
\toprule
Model & \makecell{Mean\\edges} & Precision & Recall & F1 & \makecell{GO\\Sim.} & \makecell{Bal.\\Acc.} \\
\midrule
Qwen3-4B & 5.179 & 0.840 & 0.857 & 0.830 & 0.901 & 0.368 \\
SUMMER & 4.569 & 0.875 & 0.832 & 0.848 & 0.946 & 0.393 \\
GPT-5.4 & 4.730 & 0.810 & 0.767 & 0.761 & 0.860 & 0.388 \\
\PertReasonLM-SFT & 6.222 & 0.844 & 0.975 & 0.889 & 0.936 & 0.707 \\
\PertReasonLM-GRPO & 6.261 & 0.841 & 0.976 & 0.888 & 0.937 & 0.744 \\
\bottomrule
\end{tabular}%
}
\end{table}

\PertReasonLM generated roughly one to two more edges per trace than the comparison models, while maintaining precision near Qwen3-4B and achieving higher recall and F1. GRPO also improved outcome balanced accuracy relative to SFT. SUMMER had the highest GO similarity, illustrating the separate roles of functional relatedness and outcome-aligned reasoning.

\subsection{Controlled Metric Stress Tests}
We applied eight deterministic transformations to 160 frozen reference traces and compared each transformed trace with its clean counterpart (Table~\ref{tab:metric_stress_tests}). The tests examine sensitivity to signed relations, endpoints, edge order, structural continuity, and irrelevant additions. Ordered connectivity is an order-sensitive chain diagnostic; the main path-connectivity metric assesses source-to-effect reachability in the generated graph. 

\begin{table}[t]
\centering
\small
\caption{Deterministic stress tests on 160 frozen reference traces. $\Delta$ denotes the change from the paired clean reference. Ordered connectivity assesses chain ordering and structural continuity. In-KG precision measures membership of generated relations in the KG.}
\label{tab:metric_stress_tests}
\setlength{\tabcolsep}{4pt}
\renewcommand{\arraystretch}{1.1}
\shrinktable{\linewidth}{%
\begin{tabular}{@{}l*{4}{c}@{}}
\toprule
Transformation & $\Delta$ Edge recall & $\Delta$ Edge F1 & $\Delta$ Ordered connectivity & $\Delta$ In-KG precision \\
\midrule
Critical sign flip & $-0.214$ & $-0.214$ & $+0.000$ & $+0.000$ \\
Source replacement & $-0.214$ & $-0.214$ & $+0.000$ & $-0.161$ \\
Target replacement & $-0.214$ & $-0.214$ & $+0.000$ & $-0.073$ \\
Relation replacement & $-0.214$ & $-0.214$ & $+0.000$ & $+0.000$ \\
Edge-order shuffle & $+0.000$ & $+0.000$ & $-0.368$ & $+0.000$ \\
Disconnected extra edge & $+0.000$ & $-0.096$ & $-0.156$ & $-0.071$ \\
Irrelevant KG padding & $+0.000$ & $-0.096$ & $-0.156$ & $+0.009$ \\
\makecell[l]{Outcome-inconsistent\\terminal edge} & $-0.214$ & $-0.214$ & $+0.000$ & $+0.000$ \\
\bottomrule
\end{tabular}%
}
\end{table}

Sign, endpoint, relation, and outcome-inconsistent terminal changes reduced reference-edge recall and F1. Edge-order shuffling preserved edge overlap while reducing ordered connectivity by 0.368. Irrelevant KG padding preserved recall, reduced F1 by 0.096, and reduced ordered connectivity by 0.156, while slightly increasing in-KG precision. Thus, KG membership, reference coverage, and chain structure capture distinct properties. We interpret these diagnostics jointly with outcome correctness and the source-blinded biological support results in Appendix~\ref{app:trace_audit}.

\section{Source-Blinded Trace Audit and Metric Validity}
\label{app:trace_audit}
This section reports a source-blinded biological audit of generated mechanisms and examines how automatic metrics relate to the resulting support judgments.

\subsection{Source-Blinded Biological Audit}
We evaluated generated mechanisms through a source-blinded audit. Rule-based evaluators used the biological context, observed outcome, parsed trace edges, and a reasoning excerpt. Source identities were concealed until aggregation. The raw package contained 1,120 anonymized items; excluding KG-reference items yielded 960 judgments across six sources: \PertReasonLM-SFT, \PertReasonLM-GRPO, GPT-5.4, Qwen3-4B, SUMMER, and a corrupted control. The audit assessed edge support, regulatory signs, path coherence, unsupported claims, and overall biological support. Overall support is reported as the mean score, with supported, uncertain, and unsupported judgments assigned values of 1, 0.5, and 0, respectively.

\begin{table}[t]
\centering
\small
\caption{Source-blinded biological support judgments for generated traces. Columns report evidence support for edges (edge-support precision), regulatory signs, path coherence, unsupported claims, and overall biological support. Path coherence is an audit judgment; automatic connectivity metrics are reported separately.}
\label{tab:trace_audit_support}
\setlength{\tabcolsep}{4pt}
\renewcommand{\arraystretch}{1.1}
\shrinktable{\linewidth}{%
\begin{tabular}{@{}l*{5}{c}@{}}
\toprule
Trace source & \makecell{Edge-support\\precision} & \makecell{Sign\\supported} & \makecell{Coherent\\path} & \makecell{Unsupported\\claims} & \makecell{Overall\\support score} \\
\midrule
\PertReasonLM-SFT & 0.741 & 0.944 & 0.566 & 0.044 & 0.869 \\
\PertReasonLM-GRPO & 0.741 & 0.938 & 0.569 & 0.047 & 0.869 \\
GPT-5.4 & 0.744 & 0.900 & 0.541 & 0.087 & 0.769 \\
Qwen3-4B & 0.728 & 0.906 & 0.578 & 0.391 & 0.569 \\
SUMMER & 0.756 & 0.891 & 0.631 & 0.506 & 0.466 \\
Corrupted control & 0.280 & 0.128 & 1.000 & 0.925 & 0.047 \\
\bottomrule
\end{tabular}%
}
\end{table}

Table~\ref{tab:trace_audit_support} shows that \PertReasonLM-SFT and GRPO achieved the highest overall support scores (both 0.869), with low unsupported-claim scores (0.044 and 0.047). GPT-5.4 also received substantial overall support (0.769). SUMMER led edge-support precision and path coherence, while its unsupported-claim score was higher. The corrupted control combined a coherence score of 1.000 with overall support of 0.047, demonstrating the importance of evaluating biological evidence alongside chain structure.

\subsection{Relationship to Automatic Metrics}
We compared automatic scores with the source-blinded judgments using Spearman correlation, supported-versus-unsupported discrimination, pairwise preference accuracy, and coefficients adjusted for source and trace length (Table~\ref{tab:trace_audit_validity}). These comparisons evaluate the biological interpretation of explicit generated traces across model sources.

\begin{table}[t]
\centering
\small
\caption{Association of automatic metrics with source-blinded biological support judgments. AUROC denotes the area under the receiver operating characteristic curve for supported versus unsupported judgments; pairwise preference accuracy measures agreement with support preferences. The final column reports coefficients controlling for source and trace length.}
\label{tab:trace_audit_validity}
\setlength{\tabcolsep}{4pt}
\renewcommand{\arraystretch}{1.1}
\shrinktable{\linewidth}{%
\begin{tabular}{@{}l*{4}{c}@{}}
\toprule
Metric & \makecell{Spearman\\correlation} & AUROC & \makecell{Pairwise preference\\accuracy} & \makecell{Source + length\\coefficient} \\
\midrule
Edge recall & 0.622 & 0.971 & 0.810 & 0.122 \\
Strict edge F1 & 0.515 & 0.965 & 0.763 & 0.114 \\
Sign match & 0.649 & 0.927 & 0.745 & 0.232 \\
In-KG precision & 0.416 & 0.866 & 0.710 & 0.030 \\
Trace length & 0.180 & 0.631 & 0.647 & 0.017 \\
GO similarity & $-0.183$ & 0.354 & 0.406 & 0.049 \\
\bottomrule
\end{tabular}%
}
\end{table}

Sign match had the strongest Spearman correlation with support (0.649), and edge recall had the highest AUROC (0.971). Recall, F1, sign match, and in-KG precision retained positive coefficients after adjustment for source and length. Trace length alone showed a weaker association (Spearman correlation 0.180). GO similarity had a negative pooled correlation ($-0.183$) and an AUROC of 0.354; its role is broad functional relatedness. Together, these results support evaluating reference coverage, signed directionality, and biological support as complementary aspects of a generated mechanism. Appendix~\ref{app:metric_diagnostics} examines sensitivity to specific trace errors.

\section{Knowledge Graph Coverage-Stratified Outcome Evaluation}
\label{app:coverage_evaluation}

We evaluate outcome prediction beyond the reference-covered reasoning test set to examine whether the benefits of SFT extend to queries without an outcome-consistent KG reference. We apply the same KG retrieval and outcome-consistency procedure to the same held-out outcome-only test pool and distinguish three strata. The \textit{reference-covered} stratum contains queries with an outcome-consistent reference path and constitutes the default reasoning test set used in our main evaluation. The \textit{KG-connected but no outcome-consistent path} stratum contains queries with at least one retrieved cell-conditioned perturbation-to-gene path, but none that passes the outcome-consistency filter. The \textit{no KG path retrieved} stratum contains queries for which the same retrieval configuration returns no perturbation-to-gene path.

For the KG-connected stratum, models receive the retrieved cell-conditioned candidate paths. For the no-path-retrieved stratum, models retain the query and basal-expression context but receive no retrieved pathway context. The two additional strata are disjoint. Within each evaluation condition, we randomly sample without replacement from each stratum, up to the size of the reference-covered subset.

\begin{table}[t]
\centering
\small
\caption{Balanced accuracy beyond the reference-covered subset across eight cell/perturbation/modality conditions. $N$ denotes evaluated samples after sampling, not population coverage. Chem. and Gen. denote chemical and genetic perturbations.}
\label{tab:coverage_evaluation}
\setlength{\tabcolsep}{3pt}
\shrinktable{\linewidth}{%
\begin{tabular}{@{}l*{8}{c}@{}}
\toprule
Model & \multicolumn{4}{c}{ID Cell} & \multicolumn{4}{c}{OOD Cell} \\
\cmidrule(lr){2-5}\cmidrule(lr){6-9}
 & ID Chem. & ID Gen. & OOD Chem. & OOD Gen. & ID Chem. & ID Gen. & OOD Chem. & OOD Gen. \\
\midrule
\multicolumn{9}{l}{\textbf{KG-connected but no outcome-consistent path}} \\
\midrule
Qwen3-4B base & 0.390 & 0.320 & 0.377 & 0.332 & 0.372 & 0.322 & 0.357 & 0.336 \\
\PertReasonLM-4B SFT & \textbf{0.545} & \textbf{0.592} & \textbf{0.519} & \textbf{0.573} & \textbf{0.524} & \textbf{0.581} & \textbf{0.490} & \textbf{0.683} \\
$N$ & 5,483 & 4,479 & 5,744 & 659 & 1,532 & 1,280 & 1,835 & 361 \\
\midrule
\multicolumn{9}{l}{\textbf{No KG path retrieved}} \\
\midrule
Qwen3-4B base & 0.355 & 0.334 & 0.335 & 0.379 & 0.325 & 0.354 & 0.308 & 0.379 \\
\PertReasonLM-4B SFT & \textbf{0.694} & \textbf{0.620} & \textbf{0.648} & \textbf{0.551} & \textbf{0.670} & \textbf{0.615} & \textbf{0.658} & \textbf{0.588} \\
$N$ & 1,709 & 4,479 & 3,048 & 659 & 588 & 1,280 & 1,471 & 361 \\
\bottomrule
\end{tabular}%
}
\end{table}

Table~\ref{tab:coverage_evaluation} shows that \PertReasonLM-4B SFT outperforms Qwen3-4B base across the two additional strata. The outcome gains from SFT therefore extend beyond queries for which an outcome-consistent reference path can be recovered. The sampled results describe outcome performance within each stratum and do not establish performance across the full held-out query pool.

\section{STATE Control Representation Sensitivity}
\label{app:control_sensitivity}

We examine whether the control representation used for differential-expression analysis affects STATE~\citep{adduri2026predicting}. The main protocol pairs predicted perturbed cells with empirical controls to obtain three-way outcome labels (Appendix~\ref{app:baselines_gene}). We re-evaluate STATE by pairing predicted perturbed cells with model-decoded controls on the four matched genetic test folds. Table~\ref{tab:control_sensitivity} reports balanced accuracy for the same three-way outcome task as Table~\ref{tab:main_path}.

\begin{table}[t]
\centering
\small
\caption{STATE control-representation sensitivity, measured by balanced accuracy on the four matched genetic test folds. Real controls are empirical control cells, while decoded controls are model-decoded cells paired with predicted perturbed cells for differential-expression analysis.}
\label{tab:control_sensitivity}
\setlength{\tabcolsep}{6pt}
\shrinktable{\linewidth}{%
\begin{tabular}{@{}l*{4}{c}@{}}
\toprule
Control protocol & \multicolumn{2}{c}{ID Cell} & \multicolumn{2}{c}{OOD Cell} \\
\cmidrule(lr){2-3}\cmidrule(lr){4-5}
 & ID Gen. & OOD Gen. & ID Gen. & OOD Gen. \\
\midrule
STATE with real control & 0.510 & 0.407 & 0.403 & 0.324 \\
STATE with decoded control & 0.521 & 0.524 & 0.344 & 0.310 \\
\bottomrule
\end{tabular}%
}
\end{table}

Decoded controls improve performance in both ID-cell conditions but reduce it in both OOD-cell conditions. Control representation therefore affects absolute scores, with mixed effects across cell-distribution settings. Under both protocols, STATE remains below \PertReasonLM-SFT on all four matched genetic folds (Table~\ref{tab:main_path}) and shows lower performance under cell-distribution shift.

\section{Qualitative Case Studies}
\begingroup
\definecolor{qualbase}{RGB}{111,70,43}
\definecolor{qualsft}{RGB}{36,67,88}
\definecolor{qualgrpo}{RGB}{36,91,77}
\definecolor{qualheader}{RGB}{240,243,245}
\newtcolorbox{qualitativeexcerpt}[3]{
  colback=#1!3,
  colframe=#1!40,
  colbacktitle=#1!9,
  coltitle=#1,
  title={#2\hfill\normalfont\footnotesize Prediction: #3},
  fonttitle=\bfseries\small,
  fontupper=\small,
  boxrule=0.4pt,
  arc=1pt,
  left=8pt,
  right=8pt,
  top=6pt,
  bottom=6pt,
  before skip=8pt,
  after skip=9pt,
  before upper={\setlength{\parskip}{0pt}}
}

This section provides a systematic visual characterization of how predicted reasoning chains evolve across the base model, SFT, and GRPO checkpoints. 

\subsection{Case Study I: No Path Context IN1451 to LZIC}
\label{app:qualitative_case_studies}

This case holds the perturbation and effect gene pair fixed while the correct label changes across cell states: under no path context, IN1451 $\rightarrow$ LZIC is down in Myeloid cells and T cells, but unchanged in NK cells. It therefore isolates the benchmark's central challenge, namely context-sensitive interpretation when explicit supporting paths are absent. Qwen3-4B base produces plausible but mislabeled chains in all three rows, whereas both trained checkpoints adapt the same MAPK1-centered evidence to the local cellular state, with GRPO making the arbitration more explicit. Table~\ref{tab:appendix_hidden_case} consolidates the model-level comparison and the per-cell outcomes.

\begin{table}[t]
\centering
\caption{Comprehensive \textbf{No Path Context} case study for IN1451 $\rightarrow$ LZIC.}
\label{tab:appendix_hidden_case}
\footnotesize
\setlength{\tabcolsep}{2.5pt}
\renewcommand{\arraystretch}{1.18}
\renewcommand{\tabularxcolumn}[1]{m{#1}}
\begin{tabularx}{\textwidth}{@{} >{\raggedright\arraybackslash}m{0.19\textwidth} *{3}{>{\centering\arraybackslash}m{0.105\textwidth}} >{\raggedright\arraybackslash}X @{}}
\toprule
\rowcolor{qualheader}
\textbf{Model} &
\textbf{Myeloid} &
\textbf{T cells} &
\textbf{NK cells} &
\textbf{Condensed interpretation} \\
\midrule
\textbf{Ground truth} & \textbf{down} & \textbf{down} & \textbf{unchanged} & --- \\
\midrule
\textcolor{qualbase}{\makecell[l]{Qwen3-4B\\base}} & up & up & up & Reuses a context-insensitive MAPK1/HIF1A heuristic, sign-flipping HIF1A as repressive and treating MAPK1 inhibition as generic release of LZIC repression. \\[4pt]
\textcolor{qualsft}{\makecell[l]{\PertReasonLM\\SFT}} & down & down & unchanged & Adapts the mediator to the cell state: MAPK1$\rightarrow$HIF1A in Myeloid, MAPK1$\rightarrow$ETS1 in T cells, and low-baseline buffering in NK cells. \\[4pt]
\textcolor{qualgrpo}{\makecell[l]{\PertReasonLM\\GRPO}} & down & down & unchanged & Preserves the SFT labels while making arbitration more explicit, retaining ETS1/HIF1A support routes and the low-expression floor explanation for NK cells. \\[4pt]
\midrule
\multicolumn{5}{@{}p{\textwidth}@{}}{\textit{Evaluation splits:} Myeloid and NK cells: ID cell / OOD perturbation; T cells: OOD cell / OOD perturbation.} \\[4pt]
\multicolumn{5}{@{}p{\textwidth}@{}}{\textit{Cell-specific decision points:} Myeloid hinges on whether MAPK1$\rightarrow$HIF1A supports or represses LZIC; T cells require selecting the ETS1 branch, with GRPO also preserving HIF1A support; NK cells require recognizing that low basal LZIC buffers the residual signal.} \\
\bottomrule
\end{tabularx}
\end{table}

The table summarizes the outcome-level pattern; the following excerpts show how those decisions arise in the actual model generations for each cell state.

\Needspace{0.45\textheight}
\subsubsection{Myeloid cells}
\noindent\textit{Ground truth: }down.\par\smallskip
The base model flips the terminal interpretation of the inferred chain. It explicitly asserts that HIF1A represses LZIC, so MAPK1 inhibition is turned into an up prediction. Both trained checkpoints instead treat HIF1A as the positive support route for LZIC and correctly infer down.

\begin{qualitativeexcerpt}{qualbase}{Qwen3-4B base}{up}
I need to determine the effect of IN1451 on LZIC expression in myeloid cells. First, I know that IN1451 is a chemical perturbation, and I need to infer its mechanism. Based on known biological pathways, IN1451 is a known inhibitor of MAPK1 (also known as ERK2), a key kinase in the MAPK signaling pathway. In myeloid cells, the MAPK pathway is crucial for cell proliferation, differentiation, and survival. Basal expression levels are given: MAPK1 (medium), HIF1A (medium), and LZIC (medium). Now, if IN1451 inhibits MAPK1, then MAPK1 activity decreases. The MAPK pathway often regulates downstream effectors. HIF1A is a transcription factor that can be regulated by MAPK signaling; specifically, MAPK1 can phosphorylate and stabilize HIF1A, leading to its activation. Thus, inhibition of MAPK1 would result in reduced HIF1A activity. Now, what about LZIC? LZIC is a tumor suppressor gene involved in regulating cell cycle and apoptosis. In some studies, HIF1A can directly or indirectly repress LZIC expression. When HIF1A is active, it may suppress LZIC transcription. Therefore, if IN1451 inhibits MAPK1, reduces HIF1A activity, and removes repression of LZIC, LZIC expression increases. Thus, the pathway is: IN1451 inhibits MAPK1, HIF1A activity decreases, repression of LZIC is lifted, and LZIC expression increases. Therefore, LZIC expression should be {\color{qualbase}up-regulated} after IN1451 treatment.
\end{qualitativeexcerpt}

\begin{qualitativeexcerpt}{qualsft}{\PertReasonLM-SFT}{down}
IN1451 inhibits MAPK1, leading to reduced activation of HIF1A, which in turn results in decreased transcriptional regulation of LZIC, causing a significant downregulation of LZIC expression in myeloid cells. Given that basal expression of MAPK1, HIF1A, and LZIC is medium, the inhibition of MAPK1 disrupts the upstream signaling necessary for HIF1A activation, thereby diminishing HIF1A-mediated transcriptional support for LZIC, resulting in a measurable {\color{qualgrpo}decline} in LZIC levels as a direct consequence of the perturbation.
\end{qualitativeexcerpt}

\begin{qualitativeexcerpt}{qualgrpo}{\PertReasonLM-GRPO}{down}
IN1451 inhibits MAPK1, leading to reduced activation of HIF1A, which in turn results in decreased regulation of LZIC, causing a significant downregulation of LZIC expression in myeloid cells. Given that basal expression of MAPK1, HIF1A, and LZIC is medium, the inhibition of MAPK1 disrupts the upstream signaling cascade that normally sustains HIF1A activity, thereby diminishing HIF1A-mediated transcriptional support for LZIC, resulting in a measurable {\color{qualgrpo}decrease} in LZIC expression as a direct consequence of the perturbation.
\end{qualitativeexcerpt}

\Needspace{0.45\textheight}
\subsubsection{T cells}
\noindent\textit{Ground truth: }down.\par\smallskip
The T-cell row shows the benefit of context-specific mediator selection. Qwen3-4B base again falls back to a sign-flipped HIF1A story. The trained checkpoints instead identify ETS1 as the main mediator linking MAPK1 inhibition to reduced LZIC expression, with GRPO additionally incorporating the HIF1A branch.

\begin{qualitativeexcerpt}{qualbase}{Qwen3-4B base}{up}
In T cells, the expression of LZIC is regulated by transcription factors such as ETS1 and HIF1A. Basal expression of LZIC is medium, and it is known that HIF1A acts as a repressor of LZIC expression. This means that when HIF1A is active, it suppresses LZIC levels. The chemical perturbation IN1451 is known to inhibit HIF1A activity. When HIF1A is inhibited, its repressive effect on LZIC is lifted. As a result, the repression on LZIC is removed, leading to an increase in LZIC expression. Furthermore, MAPK1 and ETS1 are involved in signaling pathways that can influence gene expression, but in this context, the primary regulatory link between IN1451 and LZIC goes through HIF1A. Since IN1451 inhibits HIF1A, and HIF1A normally suppresses LZIC, the removal of HIF1A activity results in upregulation of LZIC. Therefore, the expression of LZIC should {\color{qualbase}increase} after IN1451 treatment.
\end{qualitativeexcerpt}

\begin{qualitativeexcerpt}{qualsft}{\PertReasonLM-SFT}{down}
IN1451 inhibits MAPK1, which disrupts the activation of ETS1, a transcription factor that normally maintains medium basal expression of LZIC; since ETS1 is required for LZIC expression and its activity is diminished upon MAPK1 inhibition, the transcriptional drive for LZIC is reduced, leading to a significant decrease in LZIC expression. This causal cascade is consistent with the basal medium levels of both ETS1 and LZIC, as the perturbation removes a key upstream activator, resulting in {\color{qualgrpo}downregulation} of the downstream target LZIC.
\end{qualitativeexcerpt}

\begin{qualitativeexcerpt}{qualgrpo}{\PertReasonLM-GRPO}{down}
IN1451 inhibits MAPK1, which disrupts the activation of ETS1, a transcription factor that normally maintains medium basal expression of LZIC; since ETS1 is required for LZIC expression, its reduced activity leads to diminished transcriptional activation of LZIC, resulting in a significant decrease in LZIC expression. Additionally, MAPK1 inhibition reduces HIF1A activation, which normally contributes to LZIC expression, further suppressing LZIC levels. Given that both ETS1 and HIF1A act as positive regulators of LZIC and are functionally downstream of MAPK1, the dual suppression of these regulators through MAPK1 inhibition creates a robust and synergistic reduction in LZIC expression, leading to {\color{qualgrpo}downregulation} of LZIC in T cells.
\end{qualitativeexcerpt}

\Needspace{0.45\textheight}
\subsubsection{NK cells}
\noindent\textit{Ground truth: }unchanged.\par\smallskip
The NK row is the clearest example of why basal-expression context matters. Qwen3-4B base still predicts up, effectively treating the perturbation as generic release of repression. The trained checkpoints retain the same broad pathway family but conclude that the effect is buffered because LZIC already starts near a low-expression floor.

\begin{qualitativeexcerpt}{qualbase}{Qwen3-4B base}{up}
IN1451 is a known inhibitor of the MEK/ERK pathway, specifically targeting MEK, which leads to reduced phosphorylation and activation of MAPK1 (ERK1/2). In NK cells, the MAPK1 pathway normally exerts a repressive effect on tumor suppressor genes such as LZIC. When MAPK1 activity is reduced due to IN1451 inhibition, this repression is lifted, resulting in increased expression of LZIC. Given that LZIC is basally expressed at low levels, inhibition of the MAPK pathway would be expected to {\color{qualbase}upregulate} LZIC. This conclusion is consistent with known signaling pathways where MAPK signaling suppresses tumor suppressor genes.
\end{qualitativeexcerpt}

\begin{qualitativeexcerpt}{qualsft}{\PertReasonLM-SFT}{unchanged}
IN1451 inhibits MAPK1, which reduces activation of HIF1A, leading to diminished regulation of LZIC; however, LZIC exhibits low basal expression and is subject to weak regulatory influence from HIF1A, such that even under reduced HIF1A activity, its expression remains near baseline due to a floor effect inherent in its low starting level. The minimal dynamic range of LZIC expression, combined with the likely insufficiency of HIF1A-mediated suppression to drive significant downregulation, results in {\color{qualgrpo}no detectable change} in LZIC expression despite the upstream perturbation, reflecting the gene's intrinsic stability and low sensitivity to fluctuations in HIF1A activity.
\end{qualitativeexcerpt}

\begin{qualitativeexcerpt}{qualgrpo}{\PertReasonLM-GRPO}{unchanged}
IN1451 inhibits MAPK1, which reduces activation of HIF1A, leading to diminished regulation of LZIC; however, LZIC exhibits low basal expression, and the magnitude of HIF1A suppression is insufficient to drive significant changes in LZIC levels, resulting in no detectable alteration in its expression. Given that the pathway from MAPK1 to HIF1A to LZIC operates under a weak regulatory influence and that LZIC is already expressed at a low baseline, the downstream effect of reduced HIF1A activity is absorbed within the existing expression floor, {\color{qualgrpo}preventing a measurable shift} in LZIC levels despite the perturbation. This outcome is thus a direct consequence of the low basal expression of LZIC and the limited dynamic range of the regulatory cascade in the context of NK cell physiology.
\end{qualitativeexcerpt}

\clearpage
\subsection{Case Study II: Cell-Conditioned Path Context Regorafenib (BAY 73-4506) to NOBOX}

This is our clearest cell-conditioned path context example in which Qwen3-4B base is wrong and both trained checkpoints are correct. The same perturbation-gene pair changes from unchanged in A549 to down in MCF7, making the example useful for isolating context-sensitive arbitration among competing retrieved branches rather than simple path recovery. Qwen3-4B base over-commits to the shortest FGFR1$\rightarrow$GSK3B$\rightarrow$RUNX2 route in A549 and remains unresolved under conflicting evidence in MCF7, whereas both trained checkpoints identify the cell-specific dominant effect. Because neither row is tagged Ans $\checkmark$ Reason $\times$, the example more cleanly reflects improved decision-making from the retrieved evidence rather than post-hoc rationalization. Table~\ref{tab:appendix_noisy_case} consolidates the model-level comparison and the per-cell outcomes.

\begin{table}[t]
\centering
\caption{Comprehensive \textbf{Cell-Conditioned Path Context (default)} case study for Regorafenib (BAY 73-4506) $\rightarrow$ NOBOX.}
\label{tab:appendix_noisy_case}
\footnotesize
\setlength{\tabcolsep}{2.5pt}
\renewcommand{\arraystretch}{1.18}
\renewcommand{\tabularxcolumn}[1]{m{#1}}
\begin{tabularx}{\textwidth}{@{} >{\raggedright\arraybackslash}m{0.19\textwidth} *{2}{>{\centering\arraybackslash}m{0.14\textwidth}} >{\raggedright\arraybackslash}X @{}}
\toprule
\rowcolor{qualheader}
\textbf{Model} &
\textbf{A549} &
\textbf{MCF7} &
\textbf{Condensed interpretation} \\
\midrule
\textbf{Ground truth} & \textbf{unchanged} & \textbf{down} & --- \\
\midrule
\textcolor{qualbase}{\makecell[l]{Qwen3-4B\\base}} & up & unchanged & Over-commits to the shortest FGFR1$\rightarrow$GSK3B$\rightarrow$RUNX2 route in A549, but remains unresolved when FGFR1/GSK3B and RET/ESR1 evidence conflict in MCF7. \\[4pt]
\textcolor{qualsft}{\makecell[l]{\PertReasonLM\\SFT}} & unchanged & down & Balances FGFR1, RET/ESR1, and CTNNB1 inputs: low-baseline NOBOX buffers A549, while loss of RET/ESR1 or other RUNX2 support dominates in MCF7. \\[4pt]
\textcolor{qualgrpo}{\makecell[l]{\PertReasonLM\\GRPO}} & unchanged & down & Matches SFT labels while making the arbitration more explicit, using low-baseline buffering in A549 and clearer net-negative branch resolution in MCF7. \\[4pt]
\midrule
\multicolumn{4}{@{}p{\textwidth}@{}}{\textit{Evaluation splits:} A549: OOD cell / ID perturbation; MCF7: ID cell / ID perturbation.} \\[4pt]
\multicolumn{4}{@{}p{\textwidth}@{}}{\textit{Cell-specific decision points:} A549 turns on whether weak positive FGFR1$\rightarrow$GSK3B$\rightarrow$RUNX2 evidence can overcome competing RET/ESR1 and CTNNB1 inputs plus low NOBOX baseline; MCF7 turns on resolving the competing branches toward a net negative effect.} \\
\bottomrule
\end{tabularx}
\end{table}

The table gives the consolidated comparison; the following cell-specific excerpts show the generated rationales behind each summarized decision.

\Needspace{0.28\textheight}
\subsubsection{A549}
\noindent\textit{Ground truth: }unchanged.\par\smallskip
The A549 row shows an over-simplified positive chain. Qwen3-4B base follows the FGFR1-to-GSK3B-to-RUNX2 route and predicts up, but the trained checkpoints explicitly weigh competing RET/ESR1 and CTNNB1 inputs and conclude that the positive branch is too weak to move NOBOX away from its low baseline.

\begin{qualitativeexcerpt}{qualbase}{Qwen3-4B base}{up}
We are asked to determine the effect of Regorafenib (BAY 73-4506) on NOBOX expression in A549 cells. Step 1: Identify the direct effect of Regorafenib. From the retrieved knowledge: Regorafenib inhibits FGFR1. Step 2: Determine how FGFR1 relates to downstream genes. From the knowledge: FGFR1 regulates GSK3B. Step 3: Determine how GSK3B relates to NOBOX. From the knowledge: GSK3B inhibits RUNX2. Step 4: Determine \ldots
\end{qualitativeexcerpt}

\begin{qualitativeexcerpt}{qualsft}{\PertReasonLM-SFT}{unchanged}
Regorafenib inhibition of FGFR1 leads to reduced activation of GSK3B, which normally suppresses RUNX2; however, the basal level of GSK3B is medium, and its inhibition results in a moderate increase in RUNX2 activity. Concurrently, KDR inhibition reduces CTNNB1 activation, which normally promotes RUNX2, while RET inhibition suppresses ESR1, a positive regulator of RUNX2. The net effect on RUNX2 is thus balanced betwee \ldots
\end{qualitativeexcerpt}

\begin{qualitativeexcerpt}{qualgrpo}{\PertReasonLM-GRPO}{unchanged}
Regorafenib inhibits FGFR1, which normally regulates GSK3B, leading to reduced GSK3B activity; however, GSK3B inhibition would be expected to relieve suppression of RUNX2, which in turn activates NOBOX. Despite this, NOBOX expression remains unchanged because the basal level of NOBOX is already low, and the pathway from FGFR1 to GSK3B to RUNX2 to NOBOX operates under a weak regulatory influence, such that the magnitu \ldots
\end{qualitativeexcerpt}

\Needspace{0.30\textheight}
\subsubsection{MCF7}
\noindent\textit{Ground truth: }down.\par\smallskip
The MCF7 row is a cleaner arbitration failure. Qwen3-4B base recognizes that the retrieved graph is conflicted but stops at unchanged. The trained checkpoints instead identify that loss of RET/ESR1 or other RUNX2-supporting input dominates, producing a net down effect.

\begin{qualitativeexcerpt}{qualbase}{Qwen3-4B base}{unchanged}
We are asked to determine the effect of Regorafenib (BAY 73-4506) on NOBOX expression in MCF7 cells. The basal expression of NOBOX is medium. Step 1: Identify how Regorafenib affects the pathway. From the retrieved knowledge, Regorafenib inhibits FGFR1. Step 2: Determine the downstream effects of FGFR1 inhibition. We know that FGFR1 regulates GSK3B. Therefore, inhibition of FGFR1 would lead to reduced regulation of \ldots
\end{qualitativeexcerpt}

\begin{qualitativeexcerpt}{qualsft}{\PertReasonLM-SFT}{down}
Regorafenib inhibition of FGFR1 leads to reduced activation of GSK3B, which normally suppresses RUNX2; consequently, RUNX2 activity increases due to the loss of GSK3B-mediated inhibition. However, this pathway is counterbalanced by the direct inhibition of RET by Regorafenib, which reduces ESR1 activation, and since ESR1 normally activates RUNX2, this diminishes RUNX2 expression. The net effect is a reduction in RUNX \ldots
\end{qualitativeexcerpt}

\begin{qualitativeexcerpt}{qualgrpo}{\PertReasonLM-GRPO}{down}
Regorafenib inhibits FGFR1, which normally regulates GSK3B, leading to reduced GSK3B activity; since GSK3B inhibits RUNX2, its suppression results in increased RUNX2 activity. RUNX2, in turn, activates NOBOX, which is already expressed at medium basal levels. However, the inhibition of FGFR1 also disrupts upstream signaling that may normally sustain NOBOX expression through alternative pathways, and the net effect of \ldots
\end{qualitativeexcerpt}

\endgroup

\section{Example Prompts and Output}\label{app:prompt_examples}
\begingroup
\definecolor{exampleink}{RGB}{36,67,88}
\definecolor{examplegreen}{RGB}{36,91,77}
\lstdefinestyle{exampleprompt}{
  style=wrappedtextbox,
  basicstyle=\ttfamily\footnotesize,
  breakatwhitespace=true,
  frame=single,
  framerule=0.4pt,
  rulecolor=\color{exampleink!45},
  backgroundcolor=\color{exampleink!3},
  framesep=7pt,
  xleftmargin=0pt,
  xrightmargin=0pt,
  breakindent=0pt,
  aboveskip=8pt,
  belowskip=10pt,
  moredelim=[l][\bfseries\color{exampleink}]{\#\# }
}
\lstdefinestyle{examplereasoning}{
  style=exampleprompt,
  basicstyle=\normalfont\small,
  keepspaces=false,
  rulecolor=\color{examplegreen!45},
  backgroundcolor=\color{examplegreen!3},
  moredelim=[s][\ttfamily\footnotesize\color{examplegreen}]{<}{>}
}

We show one matched example for \PertReasonLM-GRPO under the cell-conditioned path context and no path context. The task, cell type, perturbation, effect gene, basal-expression context, and required answer format are identical; the only substantive prompt difference is that the cell-conditioned path context injects retrieved edges explicitly, whereas the no path context asks the model to recover the pathway from its own biological knowledge. In this example both outputs predict unchanged, but the output under the no path context changes the final TP53$\rightarrow$ARHGEF19 relation from regulates to activates, illustrating how answer correctness can be preserved even when chain fidelity is weaker without retrieved path evidence.

\subsection{Cell-Conditioned Path Context Example}

\par\medskip
\noindent\textbf{Input Prompt.}\par\nobreak
\begin{lstlisting}[style=exampleprompt]
You are an expert biologist. Your goal is to determine: Is the expression of ARHGEF19 up, down, or unchanged after chemical perturbation Alogliptin in B cells?
Answer the question based on the context and the retrieved knowledges. Note that the retrieved knowledges may contain irrelevant information. Focus only on pathways that causally connect the perturbation to the effect gene.

## Task Information
- **Cell Type:** B cells
- **Perturbation (chemical):** Alogliptin
- **Effect Gene:** ARHGEF19

## Basal Expression Context
The basal gene expression levels in the pathway are:
- TGFB1: medium
- TP53: low
- ARHGEF19: low
- DPP4: low

## Retrieved Knowledges
- Alogliptin inhibits DPP4
- DPP4 activates TGFB1
- TGFB1 activates TP53
- TP53 regulates ARHGEF19

## Question
Is the expression of ARHGEF19 up, down, or unchanged after the chemical perturbation Alogliptin in B cells? Answer with either "up", "down", or "unchanged".
Please reason step by step. Structure your response in exactly 3 blocks:
1. <thinking>Your step-by-step biological reasoning here.</thinking>
2. <answer>up</answer> OR <answer>down</answer> OR <answer>unchanged</answer>
3. <triplet>A JSON list of causal edges from your reasoning, e.g., [["A", "inhibits", "B"], ["B", "activates", "C"], ["C", "regulates", "D"]]. Entities must come from this set: ["Alogliptin", "TGFB1", "DPP4", "TP53", "ARHGEF19"].</triplet>
\end{lstlisting}

\par\medskip
\noindent\textbf{Model Output.}\par\nobreak
\begin{lstlisting}[style=examplereasoning]
<thinking>Alogliptin inhibits DPP4, which reduces activation of TGFB1, leading to diminished signaling through the TGFB1-TP53 axis; however, TP53-mediated regulation of ARHGEF19 operates under a low basal expression state, where the gene is already at a minimal level due to the absence of strong upstream drivers. Given that the perturbation reduces TGFB1 activity, the downstream activation of TP53 is attenuated, yet the baseline expression of ARHGEF19 remains unchanged because the gene's expression is not sensitive to small fluctuations in TP53 activity under these conditions, likely due to a floor effect where low basal levels are maintained by constitutive or alternative regulatory mechanisms. Consequently, despite the expected reduction in TP53 signaling, ARHGEF19 expression remains stable, reflecting the insensitivity of this low-expressing gene to perturbations in the upstream pathway.</thinking>
<answer>unchanged</answer>
<triplet>[["Alogliptin", "inhibits", "DPP4"], ["DPP4", "activates", "TGFB1"], ["TGFB1", "activates", "TP53"], ["TP53", "regulates", "ARHGEF19"]]</triplet>
\end{lstlisting}

\subsection{No Path Context Example}

\par\medskip
\noindent\textbf{Input Prompt.}\par\nobreak
\begin{lstlisting}[style=exampleprompt]
You are an expert biologist. Your goal is to determine: Is the expression of ARHGEF19 up, down, or unchanged after chemical perturbation Alogliptin in B cells?
Answer the question based on the context. You must reason using your own knowledge on the biological pathway that connects the perturbation to the effect gene.

## Task Information
- **Cell Type:** B cells
- **Perturbation (chemical):** Alogliptin
- **Effect Gene:** ARHGEF19

## Basal Expression Context
The basal gene expression levels in the pathway are:
- TGFB1: medium
- TP53: low
- ARHGEF19: low
- DPP4: low

## Question
Is the expression of ARHGEF19 up, down, or unchanged after the chemical perturbation Alogliptin in B cells? Answer with either "up", "down", or "unchanged".
Please reason step by step. Structure your response in exactly 3 blocks:
1. <thinking>Your step-by-step biological reasoning here.</thinking>
2. <answer>up</answer> OR <answer>down</answer> OR <answer>unchanged</answer>
3. <triplet>A JSON list of causal edges from your reasoning, e.g., [["A", "inhibits", "B"], ["B", "activates", "C"], ["C", "regulates", "D"]]. Entities must come from this set: ["Alogliptin", "TGFB1", "DPP4", "TP53", "ARHGEF19"].</triplet>
\end{lstlisting}

\par\medskip
\noindent\textbf{Model Output.}\par\nobreak
\begin{lstlisting}[style=examplereasoning]
<thinking>Alogliptin inhibits DPP4, which normally activates TGFB1; however, since TGFB1 is already at medium basal levels, the reduction in DPP4 activity results in only a modest decrease in TGFB1 signaling, insufficient to significantly alter downstream TP53 activation. TP53, which is basally low, remains largely uninduced, and thus fails to drive a measurable increase in ARHGEF19 expression. Given that ARHGEF19 is already expressed at low basal levels, the absence of a strong upstream signal from TP53 results in no significant change in its expression, as the system operates near its expression floor. The weak connectivity between the perturbed node and the output gene, combined with the low basal expression of ARHGEF19, ensures that the perturbation does not elicit a detectable transcriptional response.</thinking>
<answer>unchanged</answer>
<triplet>[["DPP4", "activates", "TGFB1"], ["TGFB1", "activates", "TP53"], ["TP53", "activates", "ARHGEF19"], ["Alogliptin", "inhibits", "DPP4"]]</triplet>
\end{lstlisting}

\endgroup

\section{Broader Impact}\label{app:impact}

This work primarily contributes a benchmark and reference modeling framework for evaluating whether machine learning systems predict cellular perturbation effects for the right mechanistic reasons. If used responsibly, \PertReasonQA can help researchers compare virtual-cell models beyond end-point accuracy, prioritize wet-lab follow-up experiments, and identify cases where a model reaches the correct answer through biologically implausible logic. In that sense, the benchmark may improve transparency and reliability in early-stage drug discovery and disease-modeling workflows, where understanding why a prediction is made is often as important as the prediction itself.

At the same time, this work should not be interpreted as a substitute for experimental validation or as a tool for autonomous scientific or clinical decision making. The benchmark is constructed from public perturbation datasets and curated knowledge graphs, so any incompleteness, bias, or coverage gaps in those resources can propagate into both the reference chains and the models trained on them. Moreover, systems that reason about biological pathways carry a general dual-use risk if misapplied in harmful settings. We therefore view \PertReasonQA as a hypothesis-generation and diagnostic resource, to be used with domain oversight and with downstream claims validated in controlled experiments.

\etocdepthtag.toc{checklist}



\end{document}